\pdfoutput=1
\documentclass[11pt]{article}

\usepackage[final]{acl}

\usepackage{times}
\usepackage{latexsym}

\usepackage{hyperref}
\usepackage{url}

\usepackage{microtype}
\usepackage{amsmath}
\usepackage{inconsolata}
\usepackage{graphicx}
\usepackage{multirow}
\usepackage{booktabs}
\usepackage{algorithm}
\usepackage{algpseudocode}
\usepackage[switch]{lineno}
\usepackage{tcolorbox}
\usepackage{wrapfig}
\usepackage{subfig}
\usepackage{times}
\usepackage{amssymb}
\usepackage{placeins}
\usepackage{pifont}
\newcommand{\cmark}{\ding{51}} % ✔
\usepackage[T1]{fontenc}
\usepackage[utf8]{inputenc}

\usepackage{microtype}

\usepackage{inconsolata}

\usepackage{graphicx}

\title{LLM-Based Multi-Agent Systems are Scalable Graph Generative Models}

\author{
    Jiarui Ji$^1$,
    Runlin Lei$^{1}$, 
    Jialing Bi$^1$, 
    {\bf Zhewei Wei}$^{1}$, 
    {\bf Xu Chen}$^{1}$, 
    {\bf Yankai Lin}$^{1}$, \\
    {\bf Xuchen Pan}$^{2}$, 
    {\bf Yaliang Li}$^{2}$, 
    {\bf Bolin Ding}$^{2}$ \\
    $^1$Gaoling School of Artificial Intelligence,  Renmin University of China, Beijing, China\\
    $^2$Alibaba Group
}

\begin{document}
\maketitle
% For modeling the dynamic graph evolution process, traditional rule-based methods struggle to capture community structures within graphs, while deep learning methods only focus on fitting training graphs. 

\begin{abstract}
  % Graph generation is a fundamental task in social, technological, and scientific analysis. 
  The structural properties of naturally arising social graphs are extensively studied to understand their evolution. Prior approaches for modeling network dynamics typically rely on rule-based models, which lack realism and generalizability, or deep learning-based models, which require large-scale training datasets.
  Social graphs, as abstract graph representations of entity-wise interactions, present an opportunity to explore network evolution mechanisms through realistic simulations of human-item interactions.
  Leveraging the pre-trained social consensus knowledge embedded in large language models (LLMs), we present GraphAgent-Generator (GAG), a novel simulation-based framework for dynamic, text-attributed social graph generation.
  GAG simulates the temporal node and edge generation processes for zero-shot social graph generation. 
  The resulting graphs exhibit adherence to seven key macroscopic network properties, achieving an 11\% improvement in microscopic graph structure metrics.
  Through the node classification benchmarking task, we validate GAG effectively captures the intricate text-structure correlations in graph generation. 
  Furthermore, GAG supports generating graphs with up to nearly 100,000 nodes or 10 million edges through large-scale LLM-based agent simulation with parallel acceleration, achieving a minimum speed-up of 90.4\%.
  The source code is available at \url{https://github.com/Ji-Cather/GraphAgent}. 
\end{abstract}

\section{Introduction}

Social graphs are mathematical structures stem from pairwise interactions between entities through nodes and edges, serving as a fundamental concept in network science. 
They are widely used to model human behaviors across various domains, including scientific research~\cite{radicchi2011citation}, economic commerce~\cite{movie_lens}, and sociology~\cite{guo2009analyzing}.
A longstanding task in network science is social graph generation. 

Given an observed social graph dataset, researchers construct models based on proposed generative mechanisms to extrapolate these observations into the growth of complex networks. 
Network science theories constitute macroscopic properties such as power-law degree distribution~\citep{clauset2009power}. In contrast, microscopic properties include graph structure metrics like degree distribution and clustering coefficient~\citep{martinkus2022spectre}. 
By comparing the macroscopic and microscopic properties of the generated and real-world graphs, researchers gain deeper insights into the mechanisms underlying graph evolution.

Existing graph generation methods can be categorized into two types: (1) Rule-based methods, which rely on preset rules to generate graphs~\citep{erdos1960evolution,barabasi1999emergence}. 
These methods are designed to capture specific macroscopic properties observed in real-world networks. 
However, the need for tailored models to capture each property complicates the integration of these methods into a unified framework. ~\citep{bergmeister2024efficient}
% Additionally, they fall short in capturing micro-level properties in dynamic graph generation~\citep{bergmeister2024efficient}. 
% 这点不一定，可能会被argue 选择的random model过于简单，所以不能capture
(2) Deep learning-based methods, which leverage self-supervised learning to capture graph structures. These methods mainly include auto-regressive methods~\citep{you2018graphrnn} and one-shot methods~\citep{vignac2023digress, bergmeister2024efficient, simonovsky2018graphvae}. 
While these techniques excel in fitting the microscopic properties of observed graphs, they face challenges when generating larger graphs beyond the size of the observed dataset~\citep{bergmeister2024efficient} and struggle to maintain macroscopic properties during the growth of complex networks. 
%  Despite their success in fitting the microscopic properties of observed graph sets, they struggle when generating graphs larger than those in the observed set~\citep{bergmeister2024efficient} and fail to capture the macro property in the graph evolution process.

The limitations of previous methods stem from their attempts to use a single model to represent all forms of entity-wise interaction processes. 
Instead of merely adhering to preset rules or fitting training data, a desirable graph generator understands how graphs are formed to generate structures that align with the underlying physical process. 
For social graphs specifically, the dynamics of human-item interactions drive the network evolution~\cite {doi:10.1073/pnas.0913149107}. Fortunately, the emergence of LLMs like LLaMA~\citep{llama3modelcard} and GPT-4~\citep{openai2023gpt4} has opened new avenues for graph generation. 
With advanced capabilities in human-like responses and decision-making capabilities, LLM-based agents can effectively simulate complex interaction processes in human activities~\citep{generative_agents}.

% In GAG, carefully designed LLM-based agents constitute the actor set, while the real-world text-attributed graph constitutes the initial item set, 由items generated by actors in simulation补充. 
In this work, we introduce GraphAgent-Generator (GAG), a novel framework for generic social graph generation. 
Our approach draws on empirically studied concepts from the social sciences, particularly bipartite models of social graphs that capture actor-item interactions.
In GAG, the actor set consists of carefully designed LLM-based agents, while the initial item set is derived from a real-world seed graph. This item set is subsequently expanded through items generated by the actors. 
We propose the S-RAG algorithm to model actor-item interactions and simulate network growth patterns originating from the seed network with parallel acceleration. 
Through continuous simulations, GAG develops diverse graphs by folding the affiliation network based on node types and edge/relation types, our main contributions are: 
% Additionally, we incorporate the parallel acceleration component to accelerate the simulation process.
% Through experiments conducted in GAG, our main contributions are: 
\textbf{(1) Graphs of Real-World Network Structures:} The generated graphs exhibit seven essential structural characteristics observed in real-world networks, including \textit{power-law degree distribution},  \textit{small-world}, \textit{shrinking diameter} and etc. Specifically, GAG surpasses the best-performing baseline by 11\% on specific evaluation metrics for graph expansion tasks.
\textbf{(2) Text-attributed Graph Generation:} 
In the node classification benchmarking task, GAG demonstrates an average improvement of 1.45 in accuracy retention compared to baseline methods. By effectively capturing the intricate relationship between textual features and graph structures, GAG generates highly realistic text-attributed graphs.
% \textbf{(2) Interpretability in Graph Generation:} 
% 这点被狠狠批判了，审稿人认为agent行为是黑盒
% LLM behavior Interpretability现在一般circuit或者auto encoder去分析，感觉和我们这边想表达的Interpretability不一样，所以造成了misunderstanding
% 感觉还是就实验结果来说比较好一些，就是可以capture intricate text-structure correlation, generate realistic text-attributed graph.
\textbf{(3) Graph Generation via Large-Scale Agent Simulation:}
The framework supports the generation of graphs across ten distinct types, accommodating up to 10 million edges or nearly 100,000 nodes through simulations with up to nearly 100,000 LLM-based agents. Additionally, the parallel acceleration accelerates the simulation with a minimum speed-up of 90.4\%.

\section{Related Work}
\textbf{Graph Generation}
As an extensively explored foundational task, existing graph generation methods fall mainly into two categories:
% Rule-based methods: These methods rely on preset rules of probabilistic priors (e.g., preferential attachment) to estimate the degree distribution of the graph. 
% They successfully generate scale-free networks but fail to capture the community structures prevalent in real-world networks.
(1) Rule-based methods, which gradually add nodes based on random ~\citep{erdos1960evolution} or preferential attachment ~\citep{barabasi1999emergence} rules; Though \citep{barabasi1999emergence} successfully models power-law degree distribution, they struggle to capture the community structures prevalent in real-world networks ~\citep{you2018graphrnn} and generate text-rich attributes.
(2) Deep Learning based methods, which aims to capture the complex and diverse real-world structures through training on network dataset, mainly fall into two categories:
Autoregressive methods ~\citep{you2018graphrnn, dai2020scalable, bergmeister2024efficient} predict edges incrementally for each new node, while one-shot methods ~\citep{simonovsky2018graphvae, de2018molgan, liu2019graph, vignac2023digress} generate entire graphs in a single step. However, these methods require large-scale training data and struggle to generate graphs outside the training distribution.
Although some progress has been made with extrapolating to out-of-distribution graphs ~\citep{bergmeister2024efficient}, the maximum size of the expanded graph is limited to 144 nodes. Moreover, they cannot generalize to new contexts beyond observation~\cite{chang2024llmsgeneratestructurallyrealistic}.

\textbf{LLM-based Human Behavior Simulation}
With LLMs demonstrating advanced capabilities in human-like responses and autonomous planning ~\citep{gao2023large}, they are increasingly recognized as a new paradigm for simulations across fields such as education ~\citep{chen2024agentverse}, social dynamics ~\citep{generative_agents}, and economics ~\citep{li-etal-2024-econagent}. 
In graph generation, \citet{de2023emergence} first explores the scale-free property of power-law distributions in LLM-based agent interactions. Subsequently, ~\citet{papachristou2024network,chang2024llmsgeneratestructurallyrealistic} examines additional social graph properties. However, these simulations often lack realism due to simplified modeling of human behavior, such as name selection, and are constrained to fewer than 100 agents. Recently, ~\citet{pan2024very} introduced AgentScope, a framework enabling large-scale multi-agent simulations for simplified human behavior, demonstrated through number-guessing games. Building on this, we have enhanced AgentScope to simulate more complex human behaviors at a large scale.

\begin{figure*}[h]
  \centering
  % \rotatebox{270}{\includegraphics[height=\linewidth]{chop_pdf/framework_3.pdf}}
  % 草图待改
  \includegraphics[width=\linewidth]{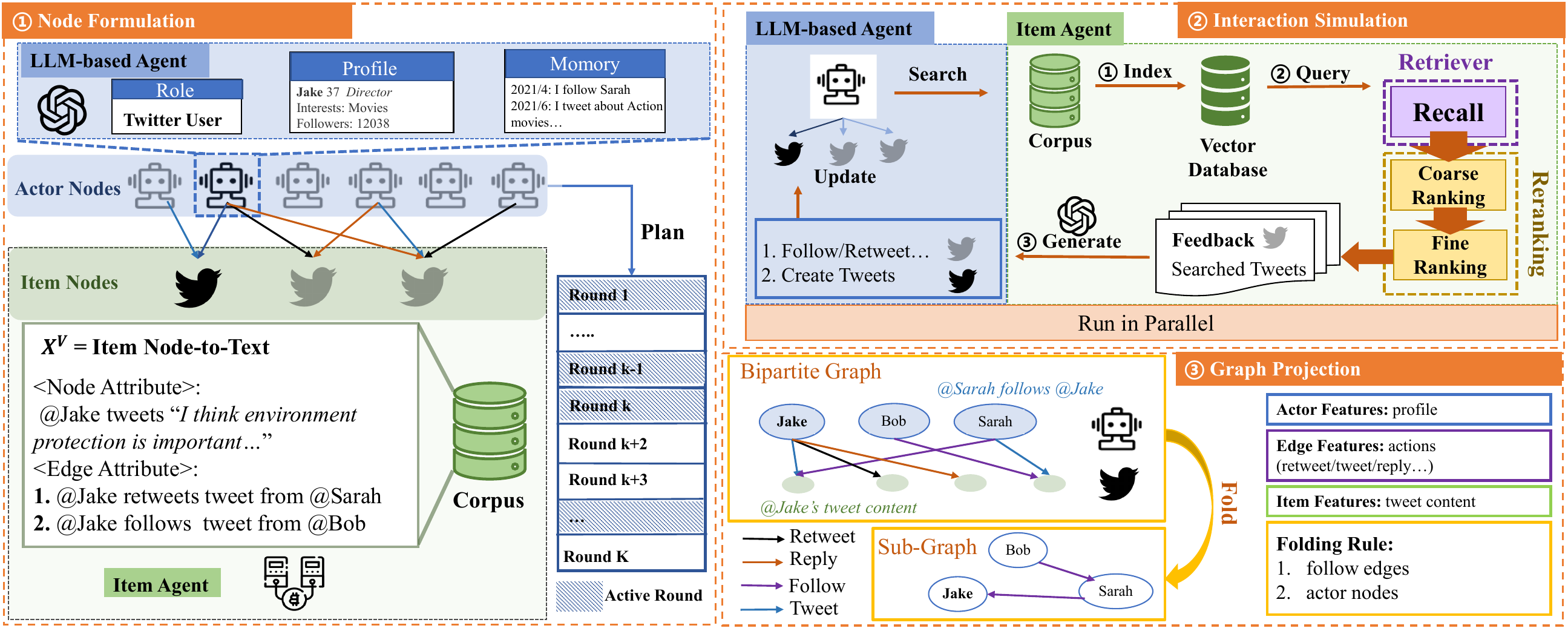}
  \caption{An illustration of the GAG Framework for generating social graphs: (1) \textbf{Node Formulation}, where actor and item node sets are initialized; actor nodes are initilized with LLM-based agents, while item nodes are managed by an item agent; (2) \textbf{Interaction Simulation}, where agents engage in pair-wise interactions within a virtual environment; (3) \textbf{Graph Projection}, where the actor-item bipartite graph is folded along specified node or edge types.}

  \label{fig:framework}
\end{figure*}
\section{The GAG Framework}

In this section, we present GAG, a versatile LLM-simulation-based framework designed for large-scale dynamic text-attributed graph generation. GAG aims to eliminate preset rules and training processes in graph generation through simulation-based methods. 
% And it is capable of generating interpretable large-scale graphs for various domains.

% \vspace{-2ex}
\subsection{Problem Setup}

% Graphs serve as mathematical structures that represent relationships and interactions among entities~\citep{20.500.11850/665023}.

% 感觉这个objective不一定准确, 应该是simulate的evolution process尽量真实,但是macro pattern和structural properties应该是会改变的
% The objective is for the generated graph \( B' \) to accurately reflect the macro patterns of $B$ while preserving its key structural properties throughout the evolution process.

In this paper, we focus on modeling the dynamic evolution of text-attributed social graphs, which include two types of entities: actors and items. The entity-wise interaction can be naturally represented as a bipartite graph \( B(\mathcal{A}, \mathcal{V}, \mathcal{E}) \), where \( \mathcal{A} \) denotes the set of actor vertices, \( \mathcal{V} \) denotes the set of item vertices, and \( \mathcal{E} \) the edges connecting them~\citep{20.500.11850/665023}. Each vertex in \( \mathcal{A} \) and \( \mathcal{V} \) is associated with textual features, represented as \( \mathcal{X}^A \) for actors and \( \mathcal{X}^V \) for items. The goal is to simulate the real-world evolution of \( B \) into a larger graph \( B' \) over time, where \( |\mathcal{A}'| \gg |\mathcal{A}| \), \( |\mathcal{V}'| \gg |\mathcal{V}| \), or \( |\mathcal{E}'| \gg |\mathcal{E}| \), while preserving graph macroscopic and microscopic properties.

To achieve this, we propose the GAG framework, which simulates actor-item interactions in human activities over \( K \) rounds of simulation. 
Leveraging the role-playing capabilities of LLMs~\citep{li2023camel, generative_agents}, we construct \( n \) LLM-based agents to form the set \( \mathcal{A} = \{a_1, a_2, \ldots, a_n\} \), simulating diverse actor entities such as authors, movie watchers, or social media users. Simultaneously, the item set \( \mathcal{V} = \{v_1, v_2, \ldots, v_m\} \) is initialized with \( m \) item entities, including papers, tweets, or movies, which is managed by an item agent.
The simulation begins with an initial text-attributed bipartite graph \( B_0 \), which serves as the seed graph. $B_0$ evolves into $B_K$ after $K$ simulation rounds.
For the \( k \)-th simulation round, $k \in [K]$, the GAG framework employs a three-step simulation workflow, as illustrated in Figure \ref{fig:framework}: \\
\textbf{(1) Node Formulation}: Given $B_{k-1}$ as input, the item set \( V_{k-1} \) and agent set \( \mathcal{A}_{k-1} \) is initialized from \( B_{k-1} \); while an actor sub-set (\( \widetilde{\mathcal{A}_k}, \widetilde{X_{k}^A}\)) of pre-set size is initialized for each simulation round. The graph updating process is as follows:
\begin{equation}
  \begin{aligned}
    \mathcal{A}_{k} &= \mathcal{A}_{ k-1} \cup \widetilde{\mathcal{A}_k},\\ 
    X_k^A &= X_{k-1}^A \cup \widetilde{X_k^A}.
  \end{aligned}  
\end{equation}
\textbf{(2) Interaction Simulation}:
Agents engage in pair-wise interactions within a virtual environment. Each interaction process generate an edge sub-set $\widetilde{E_k}$, and an item sub-set ($\widetilde{V_k}, \widetilde{X_k^V}$). The graph updating process is as follows:
\begin{equation}
  \begin{aligned}
    \mathcal{V}_{k} &= \mathcal{V}_{ k-1} \cup \widetilde{\mathcal{V}_{ k}},\\ 
    X_{k}^V &= X_{k-1}^V \cup \widetilde{X_{k}^V},\\
    \mathcal{E}_{k} &= \mathcal{E}_{k-1} \cup \widetilde{\mathcal{E}_{k}}.
  \end{aligned}  
\end{equation}
\textbf{(3) Graph Projection}: The resulting graph $B_k$ is folded along specific node types or edge types, deriving various unipartite or multipartite graphs for subsequent analysis.
% Each simulation round corresponds to a fixed real-world duration, typically set to five days or one month, depending on the specific application.  

% Additionally, we implement a \textbf{Parallel Acceleration} framework to enhance the efficiency of the simulation process.

\subsection{Node Formulation}
The underlying idea behind GAG is that in social graphs there are two types of entities — actors and items — that are related by pairwise interactions. Given \( B_{k-1} \) as input, to facilitate execution of the $k$-th simulation round, we first initialize these two node types respectively.

\paragraph{Item Node} We first collect the text feature representation of item nodes in \( B_{k-1} \). Specifically, we denote the \( j \)-th item node as \( v_{j,k-1} \), the textual feature associated as \( x_{j,k-1}^V \). To construct \( x_{j,k-1}^V \), we leverage an item template that maps each item node to its corresponding real-world textual representation~\cite{taglas}. This representation includes the node's textual features, with optional incorporation of 1-hop actor-item edge attributes.
For example, a paper entity is described using node textual features such as its title, topic, abstract, and neighboring edges, such as the authors writing the paper. 
Details of the item prompt template are provided in Appendix~\ref{appendix:item_node}.

\paragraph{Actor Node}
The textual features associated with actor nodes in \( B_{k-1} \) are denoted as \( X_{k-1}^A \).  
For the \( k \)-th simulation round, LLMs are prompted to generate a pre-set number of role-playing synthetic profiles, denoted as \( \widetilde{X_{k}^A} \). The number is fixed or proportional to the actor set size.
These profiles capture various aspects of human personal information, such as research interests, institutional affiliations, and social graph connections. The profile prompt template is detailed in Appendix~\ref{appendix:agent_formulation}.   
Consequently, the textual features of actor nodes in \( B_k \) are obtained as: $X_k^A = \{X_{k-1}^A \cup \widetilde{X_{k}^A}\}$, where $X_k^A = \{x_{i,k}^A, i \in [n]\}$.
Given $X_k^A$, GAG leverages LLM-based agents to instantiate actor nodes. The textual feature \( x_{i,k}^A \) forms the characterized profile for \( a_{i,k} \). Collectively, these LLM-based agents form the actor node set \( A_k \) of the bipartite graph.

To enable adaptive learning from past interactions, each LLM-based agent $a_{i,k}$ is equipped with a memory component that records its activity history. We organize the memory using reflection~\citep{shinn2023reflexion} and summarization techniques.  
Moreover, we adopt the action state (active or idle) to control each agent's interaction frequency. In real-world scenarios, human activity often follows a Pareto distribution~\citep{guo2009analyzing}, where approximately 20\% of users account for 80\% of the total activity.
We employ two approaches to determine the action state of each agent: 1. A fixed number of \( A_k \) is randomly sampled as active agents. 2. We first label the top 20\% agents as \textit{core}, while the remaining agents as \textit{regular} based on action history. These labels, combined with action history, are input to the LLM to determine each actor's action state each round.

\subsection{Interaction Simulation}

% actor nodes 采样
% 每一个actor node,和 item nodes 采样的subgraph交互
% 在k-th轮中,交互过程是并行的. 每一个actor node形成 edge set $\widetilde{E_{i,k}}$,以及 item node set $\widetilde{V_{i,k}}$. 因而第k-th轮将并行的输出结果合并,可以得到 edge set $\widetilde{E_{i,k}}$,以及 item node set $\widetilde{V_{i,k}}$. 

Our motivating example is the social networks that emerge from search engine queries~\cite{stoc_affiliation}. These queries enable actors to filter partial observations from the entire item set.
We first simulate this interaction process in a virtual environment, forming the edge set \( \widetilde{E_{k}} \) and a new item set \( \widetilde{V_{k}} \) for each simulation round.
To enhance efficiency, we further optimize the simulation process with parallel acceleration.

\paragraph{Interaction Process}

In real-world scenarios, humans rely on search engines to obtain efficient and targeted environment observations~\citep{10.1093/cercor/bhaa129}. Inspired by this, we propose the Simulation-Oriented Retrieval Augmented Generation (S-RAG) framework. For the \( k \)-th simulation round, active actor $a_{i,k}$ is provided with an item subset as environment observation, denoted as $O_{i,k}, O_{i,k} \subseteq V_{k-1}$. Following traditional RAG~\citep{Cuconasu_2024}, S-RAG is divided into three processes as shown in Algorithm~\ref{alg:s-rag}: 

% For the \( k \)-th simulation round, each actor \( a_{i,k} \) ideally requires access to the entire item description set \( X_{i,k}^V \) and selects a subset \( \widetilde{O_{i,k}} \) as its desired item observation. However, as \( |X_{i,k}^V| \) grows significantly over simulation rounds, the computational cost of providing full access becomes prohibitive.  
% Fortunately, humans rely on search engines to obtain efficient and targeted environment observations in real-world scenarios~\citep{10.1093/cercor/bhaa129}. 
% Inspired by this, we propose the Simulation-Oriented Retrival Augmented Generation (S-RAG) framework.
% When $a_{i,k}$ decides to act, it is provided with an item subset as environment observation, denoted as $O_{i,k}, O_{i,k} \subseteq X_{i,k}^V$.
% Our objective is to ensure that \( O_{i,k} \), determined by the retrieval process, closely aligns with the desired observation \( \widetilde{O_{i,k}} \).  
% Following traditional RAG~\citep{Cuconasu_2024}, S-RAG is divided into three processes as shown in Algorithm~\ref{alg:s-rag}: 

\begin{algorithm}
\caption{S-RAG for the $k$-th simulation round.}
\label{alg:s-rag}
\begin{algorithmic}[1]
\Require Item Set $V_{k-1}$, Large Language Model$\operatorname{LLM}$.
\State $\widetilde{V_{k}} = \emptyset, \widetilde{\mathcal{E}_{k}} = \emptyset,\widetilde{X_{k}^V} = \emptyset,$
\For{$i \in [n]$}
    \If{$a_{i,k}$ is active} 
        $Q_{i,k} = \operatorname{LLM}(a_{i,k} \mid memory)$,
    \Else{}
        continue
    \EndIf
    \State $O_{i,k} = \emptyset$,
    \For{$q \in Q_{i,k}$}
        % \State $\vec{v} = \operatorname{encoder}(q)$,
        \State $O_{i,k,q} = $ \Call{Recall}{$q, V_{k-1}$},
        \State $O_{i,k,q}= $ \Call{ReRanking}{$O_{i,k,q}, a_{i,k}$},
        \State $O_{i,k} = O_{i,k} \cup O_{i,k,q}$,
    \EndFor
    \State $\widetilde{V_{i,k}}, \widetilde{\mathcal{E}_{i,k}}, \widetilde{X_{i, k}^V}, = \operatorname{LLM}(a_{i,k} \mid O_{i,k})$,
    \State $\widetilde{V_{k}} = \widetilde{V_{k}}\cup \widetilde{V_{i,k}}$,
    \State $\widetilde{X_{k}^V} = \widetilde{X_{k}^V} \cup \widetilde{X_{i,k}^V}$,
    \State $\widetilde{\mathcal{E}_{k}} = \widetilde{\mathcal{E}_{k}} \cup \mathcal{E}_{i,k}$,
  \EndFor
  \\
\Return $\widetilde{V_{k}}, \widetilde{X_{k}^V} ,\widetilde{\mathcal{E}_{k}}$.
% \State $\mathcal{E_k} = \mathcal{E_k} \cup \widetilde{\mathcal{E}_{k}}$
\end{algorithmic}
\end{algorithm}

% (1) Index Process: conducted upon corpus refreshment; (2) Query Process: conducted each time in response to incoming queries; (3) Generation Process: personalized text generation based on the retrieved content.
\textbf{(1) Index Process:}
Given the item textual features \( X_{k-1}^V = \{x_{j, k-1}^V\}, j \in [m]\), where each \( x_{j,k-1}^V\) is stored as a text document.
We first convert these textual features into a set of embedding vectors \( E_{k-1}^V \) using an embedding model \( \operatorname{encoder}(\cdot) \). The process involves transforming each \( x_{j,k-1}^V \) into a \( d \)-dimensional embedding:  $e_{j,k-1}^V = \operatorname{encoder}(x_{j,k-1}^V) \in \mathbb{R}^d, j \in [m].$ We store these vectors in a vector database~\cite{douze2024faisslibrary}, managed by the item agent. 
This process constructs an environment for actors, thereby providing them with item observations. 

 \textbf{(2) Query Process:}  
For actor node $a_{i,k}$ in an active state, it can freely access environmental information. To obtain the most relevant items, the actor first reflects on its memory, which serves as input to the \(\operatorname{LLM}\) to create a query set $Q_{i,k}$, which contains descriptive keywords.
For each query \( q \in Q_{i,k} \), we first convert \( q \) into an embedding vector: $e_{q} = \operatorname{encoder}\left(q\right)$.
Next, we specify the desired number of retrieved feedbacks as $N_{r}$ and retrieve top $N_{r}$ items, denoted as \( O_{i,k,q} \).
The retrieved feedbacks for all queries collectively form the observation: 
\( O_{i,k} = \bigcup_{q \in Q_{i,k}} O_{i,k,q}.\)
Specifically, the retrieving process of \( O_{i,k,q} \) has two stages:

1. In the Recall stage, we filter out the top $N_{r}$ items by measuring the embedding similarity between $X_{k-1}^V$ and $q$:
\begin{align*}
  & O_{i,k,q} =  \operatorname{topN_r(q)}_{v_{j,k-1} \in V_{k-1}} \operatorname{Sim}\left(q, x_{j,k-1}^V\right) , \\
  & \operatorname{Sim}\left(q, x_{j,k-1}^V\right) = \frac{e_q \cdot e_{j,k-1}^V}{\left\|e_q\right\| \cdot\left\|e_{j,k-1}^V\right\|}, j \in [m].
 \end{align*} 

2. In the ReRanking stage, we refine and organize \(O_{i,k,q}\) according to the attribute of the active actor, which is divided into two phases: 
(1) Coarse Ranking: Items are reordered to prioritize those created by \textit{core} actors.
(2) Fine Ranking: Items are further reordered based on the actor's personal preferences. For example, for author-actor with expertise in AI, items focused on AI are prioritized in \(O_{i,k,q}\). 
The ReRanking hyperparameters are detailed in Appendix \ref{appendix:pairwise_interaction}.

\textbf{(3) Generation Process:}
In real-world scenarios, based on feedback from search engine queries, the actor acts according to the feedback. For example, in the context of author-paper interaction, the authors may generate a new paper and reference searched papers based on their perception of the environment.
To mimic this process, we instruct \( a_{i,k} \) using an action template to perform various actions. Each action forms an item-actor edge, where the action type determines the edge label. The action types for different simulation scenarios are listed in Appendix~\ref{appendix:pairwise_interaction}.  
For creation action, the actor \( a_{i,k} \) additionally creates a new item node during the interaction, denoted as \( \widetilde{x_{i,k}^V} \) (textual feature) and \( \widetilde{v_{i,k}} \) (item node).  
Moreover, the memory of active actor nodes is updated simultaneously with the action history, further refining their perception of the environment. Across all active agents, the interaction edges collectively form \( \widetilde{\mathcal{E}_{i,k}} \), while the created item nodes form \( \widetilde{\mathcal{V}_{i,k}} \) and \( \widetilde{X_{i,k}^V} \). 
As a result, the bipartite graph is updated accordingly.

% :  
% \begin{align*}
%   \mathcal{V}_{k} &= \mathcal{V}_{ k-1} \cup \widetilde{\mathcal{V}_{ k}},\\ 
%   X_{k}^V &= X_{k-1}^V \cup \widetilde{X_{k}^V},\\
%   \mathcal{E}_{k} &= \mathcal{E}_{k-1} \cup \widetilde{\mathcal{E}_{k}}
% \end{align*}

\paragraph{Parallel Acceleration}
The S-RAG enables the modeling of actor-item interaction processes in real-world scenarios. However, there remains technical barriers in supporting interaction simulation at the scale of $n = 1e^5$. 
Additionally, we note that the inference time of LLMs is substantial, leading to prolonged IO wait times for idle LLM-based actor agents. Various solutions have been proposed to address this issue, such as async~\citep{langchain} and actor architecture~\citep{gao2024agentscope}. We adopt the parallel processing technique~\citep{gao2024agentscope}. As highlighted by~\citep{clauset2004finding}, network structures often display tightly connected communities with loosely connected inter-community links. 
In GAG, we categorize agents into distinct groups based on strong internal interactions and weaker inter-group interactions. Specifically, each active actor agent form a group with the item agent. These groups run in parallel on CPU cores with $P$ ports. The implementation details in Appendix~\ref{Appendix:nested_actor}.

\subsection{Graph Projection}

$B_0$ progressively evolves into $B_K$ after $K$ rounds of interaction simulations.
For bipartite graphs, different projected sub-graphs are folded based on node and edge types. The sub-graph, denoted as $G(\mathcal{V}^s,\mathcal{E}^s)$, evolves via the evolution of $B$. Textual features associated with $\mathcal{V}^s$ is represented as \( \mathcal{X}^{Vs} \).
Following established folding rules in network science research, the sub-graphs embody different semantic interpretations. For instance, in an author-paper bipartite graph, selecting paper nodes and deriving paper-paper citation edges—i.e., hop-2 edges (paper-author-paper) in the bipartite graph—yields a paper citation network. Folding rules for different domains are provided in Appendix~\ref{Appendix:graph_projection}.

\section{Experiment}

In network science, there has long been an interest in graph structures that emerge within scientific, technological, and sociological contexts \citep{shrinking_diameter}. 
To evaluate our framework, we simulate graph generation across three representative domains:
(1) \textbf{Scientific Context} (SC): 
This domain models the dissemination of ideas, theories, and results in science. The simulation involves an author actor set interacting with a paper set, producing citation, bibliographic coupling, co-citation, author citation, and co-authorship networks~\citep{garfield2000web}. Simulation terminates at the citation network reaching 1e4 nodes.
(2) \textbf{Technological Context} (TC): This domain models customer-product interactions in digital commerce. The simulation involves a user actor set interacting with a movie set, producing movie rating and user projection networks~\citep{zhou2007bipartite}. Simulation terminates at the movie rating network reaching 1e5 edges.
(3) \textbf{Socialogical Context} (SoC): 
This domain models interpersonal communications in online social-media platform. The simulation involves an user actor set interact with tweet item set on platforms like Twitter, producing follow, friend, and action networks~\citep{de2013anatomy}. In addition to graph expansion, GAG can also generate graphs from scratch using LLM-generated graph textual features, eliminating the need for external data collection. We employ this method in SoC. Simulation terminates after 5 rounds.

\textbf{Evaluation Protocol}
To evaluate the effectiveness of the GAG, we assess three key aspects:
First, we compare the generated graph structures with real-world networks at both macro and micro scales.
Next, we evaluate the effectiveness of graph textual features with the GNN benchmarking task.
Finally, we assess the scalability of GAG in terms of generation scale and efficiency.
Details on the evaluation hyperparameters and metrics are provided in Appendix~\ref{appendix:evaluation_metrics}.

\subsection{Graph Structure Alignment}
In this paper, we investigate the generated graph structures from both macro and micro perspectives:
At the macro level, we examine the graph structure dynamics in the graph evolution and align our observations with established network science theories.
At the micro level, we compare GAG to existing graph generation models in capturing micro graph structural characteristics.

\textbf{Macro-Level Evaluation}
For macro-level structural characteristic alignment, we examine three structural characteristics observed in real-world networks~\citep{albert2002statistical}: \textit{power-law distribution, small-world phenomenon} and \textit{shrinking diameter}. Four additional structural characteristics are detailed in Appendix \ref{Appendix:macro_level_evaluation}. 
\begin{figure}[htbp]
  \centering
  \includegraphics[width=\linewidth]{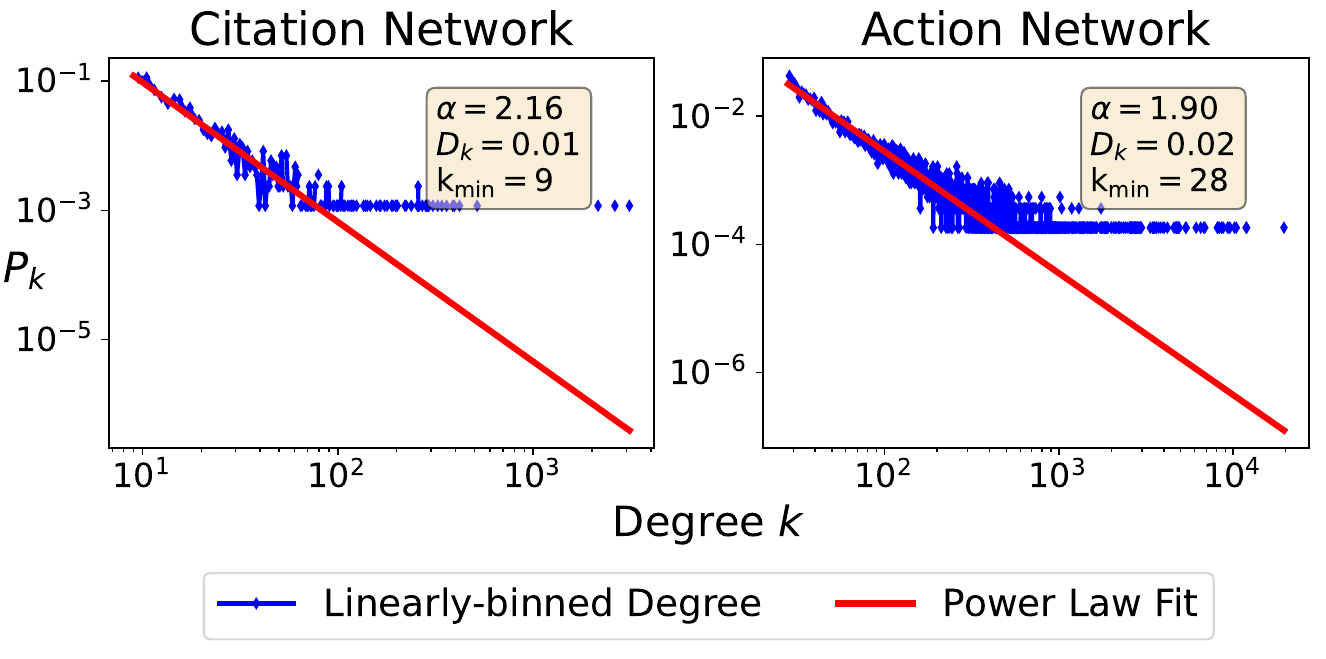}
  \caption{The \textit{power-law} Distribution of degrees in generated graphs. The degree $k$ is plotted against the probability density function $P_k$ on a log-log scale, where $\alpha$ denotes the exponent parameter, 
  $k_{min}$ represents the cut-off $k$~\citep{alstott2014powerlaw}.}
  \label{fig:power_law}
\end{figure}

\textbf{(1) Power-law Distribution:} 
The degree distribution of scale-free networks often follows a \textit{power-law distribution}~\citep{barabasi1999emergence}, 
which is commonly observed in citation networks and social networks. 
In simulations with GAG, the generated citation, author-citation, and action networks also exhibit this characteristic. We adhere to the established criteria for evaluating whether the network degree distribution follows a power law: \( D_k < 0.1 \)~\citep{alstott2014powerlaw}. As shown in Figure \ref{fig:power_law}, the degree distribution of these networks follow a \textit{power-law} distribution with exponent parameter~\citep{clauset2009power} $\alpha \in [1.90, 2.16]$.

\textbf{(2) Small World Phenomenon:}
Real-world networks exhibit a \textit{small world} phenomenon~\citep{mislove2007measurement, watts1998collective}, characterized by a small diameter and a high clustering coefficient. Table \ref{table:clustering} compares the $\bar{cc}$ of the generated graphs with that of the random graphs with consistent average degree: Erdős-Rényi~\citep{erdos1960evolution} and Barabási-Albert graphs~\citep{barabasi1999emergence}. The results indicate that generated social graphs (i.e., follow, friend, and action networks), exhibit a significantly higher $\bar{cc}$ than those of the random graphs, confirming these networks exhibit small-world characteristics.

\begin{table}[htbp]
  \centering
  \caption{$\bar{cc}$ of the generated networks, and the ratio to Erdös-Rényi and Barabási-Albert graph model. A dash (—) signifies that $\bar{cc} = 0$ for the graph model.}
  \label{table:clustering}
  \resizebox{\linewidth}{!}{%
  \begin{tabular}{lll|ll}
    \toprule
  & \multicolumn{2}{c}{Graph Scale} &\multicolumn{2}{c}{\textbf{Ratio to Random Graphs}}  \\
  \cmidrule(lr){2-3} \cmidrule(lr){4-5} 
  & $|\mathcal{V}^s|$ & \( |\mathcal{E}^s| \) & Erdös-Rényi &  Barabási-Albert \\
  \midrule
  Paper Citation & 1.14e+04 & 3.63e+04  & 301.08 &  —  \\
  Bib-Coupling & 1.09e+04 & 1.22e+07 & 7.46 & 4.40 \\
  Co-Citation & 3.93e+03 & 3.27e+04 & 275.97 & 44.87 \\
  Author Citation & 5.01e+03 & 2.41e+05 & 39.82  & 11.19 \\
  Co-Authorship & 5.01e+03 & 2.08e+04 & 234.81 & 17.59 \\
  Action & 9.97e+04 & 9.07e+05  & 784.93 & 73.97 \\
  Follow & 9.96e+04& 1.53e+06 & 3961.83 & 443.80 \\
  Friend & 9.96e+04& 5.01e+05 & 19768.58 & 1391.47 \\
  Movie Rating & 4.17e+03& 3.25e+04 & 0.00 & 0.00 \\
  User Projection & 3.91e+03& 9.04e+05 & 5.78 & 2.82 \\
  \bottomrule
  \end{tabular}  
  }
\end{table}

% Additionally, as shown in Table \ref{table:graph_structure}, the generated networks exhibit a small diameter, \( D_e \in [1.17, 11.66] \), consistent with the \textit{six degrees of separation phenomenon} observed in real-world networks~\citep{shrinking_diameter, BRODER2000309}.
% The high $\bar{cc}$, combined with small $D_{e}$, confirms these networks exhibit small-world characteristics.

\begin{figure}[htbp]
  \centering
  \subfloat[w.ReRanking]   % 第一张子图的下标（注意：注释要写在[]中括号内）
    {
        \centering
        \label{fig:shrinking_all_t}
        \includegraphics[width=0.5\linewidth]{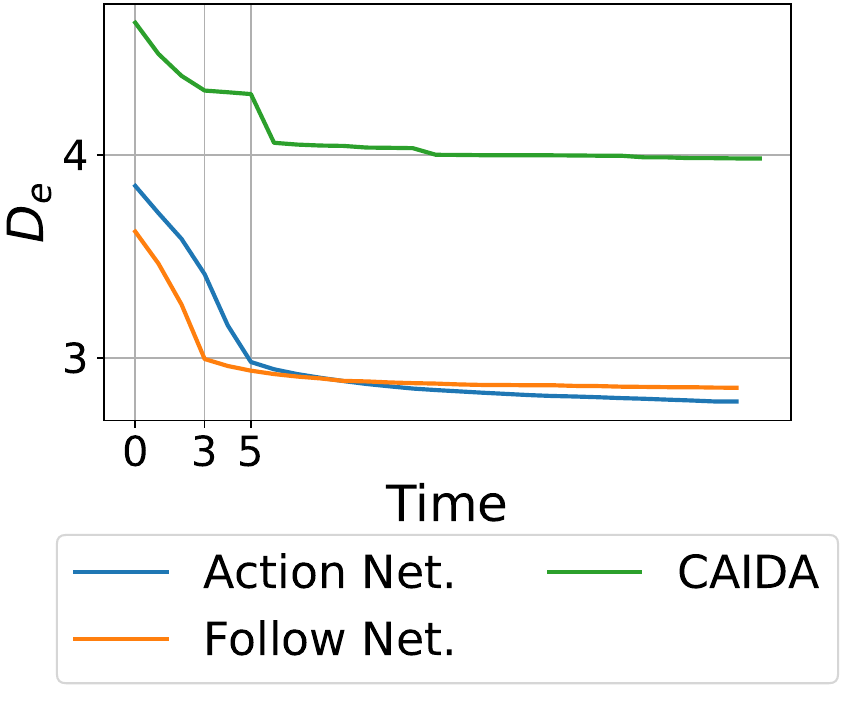}
    }
  \subfloat[w.o. ReRanking]   % 第一张子图的下标（注意：注释要写在[]中括号内）
    {
        \centering
        \label{fig:shrinking_all_f}
        \includegraphics[width=0.5\linewidth]{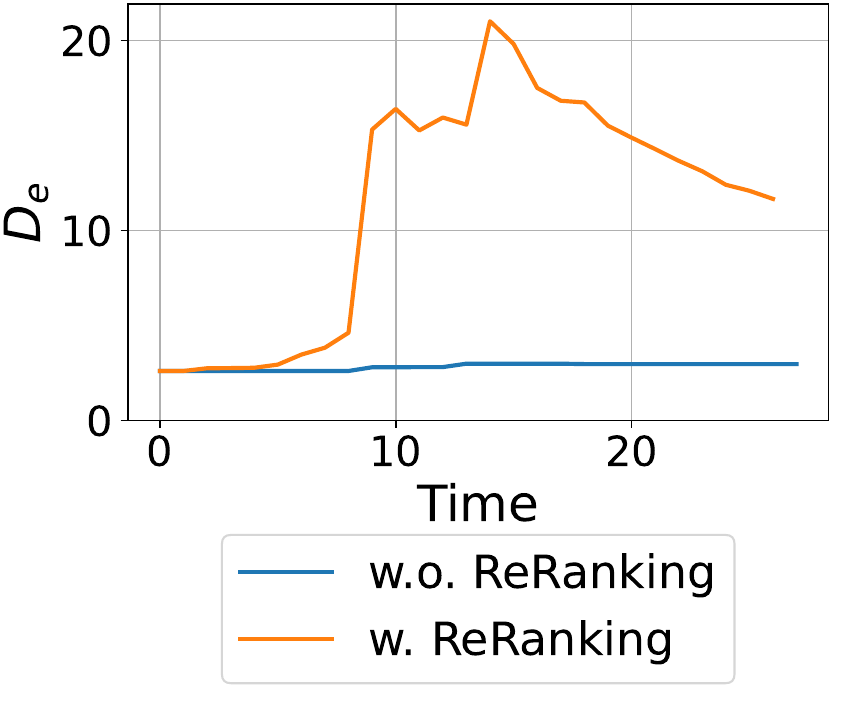}
    }
  \caption{The \textit{Shrinking Diameter} phenomenon simulated by GAG; The left figure demonstrates that as the graph evolves, $D_{e}$ gradually decreases in action and follows networks; The right figure presents an ablation experiment of the ReRanking, demonstrating its effect on $D_e$ in friend network.}
  \label{fig:shrinking_diameter_all}
\end{figure}

\textbf{(3) Shrinking Diameter:}
The \textit{shrinking diameter} is a notable phenomenon in social graphs~\citep{shrinking_diameter}, with $D_e$ decreases as the network evolves over time. 
We construct SoC with $N = 7000$, and investigate the graph evolution processes of follow, friend, and action networks for 30 simulation rounds.
We calculate the effective diameter $D_{e}$ for both the generated graphs and real-world network: CAIDA\footnote{https://sparse.tamu.edu/SNAP/as-caida}. As shown in Figure \ref{fig:shrinking_all_t}, $D_{e}$ decreases at a slow pace, identical to the trend observed in \citep{shrinking_diameter} and CAIDA.
To explain \textit{shrinking diameter}, Forest Fire model~\citep{shrinking_diameter} employs a modified preferential attachment mechanism, referred to as community-guided attachment.
In GAG, the ReRanking process enhances personalized recommendations.
We conduct an ablation experiment to assess the effect of the ReRanking. As shown in Figure~\ref{fig:shrinking_all_f}, we observe an initial increase in $D_{e}$ followed by a rapid decline. 
Notably, upon removing the ReRanking, the $D_{e}$ trends upwards from 2.6 to 2.96, indicating that ReRanking fosters community-guided attachment.
% , promoting similar community-guided attachment dynamics among agents
% To assess its impact on graph structure, 

\begin{table*}[t]
  \centering
  \caption{Comparison with existing graph generation models for graph expansion task. For GRAN and GraphMaker, the generated graphs fail to converge to a \textit{power-law} distribution.}
  \label{table:compare_structure}
  \resizebox{.88\textwidth}{!}{%
\begin{tabular}{lrrrrrrrr}
  \toprule
  &  MMD.D$\downarrow$ &  MMD.C$\downarrow$ &   MMD.S$\downarrow$ &  MMD.O$\downarrow$ &
  $D_{k}$ & $\alpha$ &  Valid$\uparrow$  &GEM\\
  \midrule
  CiteSeer     &    - &  - &      - &   - &   $0.06_{\pm 0.0}$ &   $2.38_{\pm 0.0}$ &    1.0 &  - \\
  \midrule
  Erdös-Rényi        &        0.26 &         1.41 &         0.56 &       1.41 &   $0.1_{\pm 0.01}$ &  $3.72_{\pm 0.13}$ &    0.0 &  0.34 \\
  Barabási-Albert    &        0.20 &         1.41 &         0.26 &       \textbf{1.02} &  $0.04_{\pm 0.01}$ &  $2.38_{\pm 0.04}$ &    \textbf{1.0} &  0.42 \\
  Small-World &        0.72 &         1.36 &         0.59 &       1.41 &  $0.42_{\pm 0.01}$ &   $2.03_{\pm 0.0}$ &    0.0 &  0.32 \\
  BiGG          &        0.63 &         1.13 &         0.65 &       1.23 &  $0.27_{\pm 0.01}$ &  $1.69_{\pm 0.01}$ &    0.0 &  0.33 \\
  GRAN           &        0.36 &         0.55 &         0.72 &       1.41 &    - &  $4.16_{\pm 0.39}$ &    0.0 &  0.36 \\
  BwR   &        0.49 &         1.41 &         0.66 &       1.41 &  $0.07_{\pm 0.09}$ &  $4.46_{\pm 0.02}$ &    0.0 &  0.32 \\
  GraphMaker &        0.47 &         1.41 &         0.83 &       1.41 &    - &    - &    0.0 &  0.22 \\
  L-PPGN            &        0.76 &         1.19 &         0.78 &       1.05 &  $0.39_{\pm 0.03}$ &  $1.36_{\pm 0.02}$ &    0.0 &  0.33 \\
  GAG          &   \textbf{0.16} &   \textbf{0.19} &         \textbf{0.32} &       \textbf{1.02} &  $0.08_{\pm 0.01}$ &  $2.37_{\pm 0.03}$ &    \textbf{1.0} &  \textbf{0.47} \\
  \bottomrule
  \end{tabular}
  }
\end{table*}
\textbf{Micro-Level Evaluation}
For micro-level graph structure alignment, 
we compare GAG against existing models for large-scale graph generation. Specifically, we partition the CiteSeer network~\citep{CiteGraph_Sen_08} into \( G_{<t} \) and \( G_{>t} \) based on time \( t \). 
Small subgraphs from \( G_{<t} \) are expanded and compared to the larger subgraphs of \( G_{>t} \), evaluating the model capabilities in modeling graph evolution. 
We define the Valid metric to measure the proportion of valid expanded power-law graphs and the GEM metric to evaluate the overall graph structure effectiveness. 
Details of datasets and metrics are provided in Appendix~\ref{appendix:micro_level_evaluation}.
As shown in Table \ref{table:compare_structure}, apart from the Barabási-Albert model, the graphs generated by GAG also adhere to a \textit{power-law} distribution with $\alpha = 2.39$. 
In contrast, deep-learning-based models tend to overfit the seed graph and fail to generate graphs that accurately follow the \textit{power-law} distribution. 
Surprisingly, GAG outperforms most baseline models in MMD metrics, which is notable given that no graph structure constraints are imposed on the LLM-based agents during edge creation.
Regarding GEM, GAG surpasses the best-performing baseline by 11\%. This suggests that GAG effectively simulates human behavior patterns, generating graphs that closely resemble the structural characteristics of real-world networks.

\subsection{Textual Feature Alignment}

\begin{table}[ht]
  \centering
  \caption{Benchmarking different graph generation models on the node classification task.}
  % Graph generative model for benchmarking graph neural network
  \label{tab:performance_comparison_nc}
  \resizebox{\linewidth}{!}{%
  \begin{tabular}{lllll}
  \toprule
  \multirow{2}{*}{} & \multicolumn{4}{c}{$\Delta{ACC} \downarrow$} \\
    \cmidrule{2-5}
  & GAT & GCN & GCNII & GraphSage \\
  \midrule
SF.random & $13.4_{\pm 3.5}$ & $15.0_{\pm 2.0}$ & $9.6_{\pm 0.9}$ & $7.0_{\pm 0.9}$ \\
F.random & $24.1_{\pm 2.8}$ & $23.2_{\pm 2.3}$ & $9.3_{\pm 1.4}$ & $7.1_{\pm 1.7}$ \\
S.random & $18.9_{\pm 2.5}$ & $21.8_{\pm 2.0}$ & $2.2_{\pm 1.6}$ & $3.2_{\pm 1.8}$ \\
BiGG.L & $39.4_{\pm 4.5}$ & $34.1_{\pm 6.6}$ & $4.7_{\pm 4.4}$ & $3.4_{\pm 3.5}$ \\
GRAN.L & $5.3_{\pm 3.6}$ & $6.1_{\pm 5.1}$ & $3.5_{\pm 3.4}$ & $4.4_{\pm 2.5}$ \\
BwR.L & $35.3_{\pm 2.9}$ & $38.5_{\pm 4.7}$ & $4.8_{\pm 4.5}$ & $7.6_{\pm 4.3}$ \\
GraphMaker.L & $4.0_{\pm 3.5}$ & \textbf{2.8}$_{\pm 4.0}$ & $3.6_{\pm 4.2}$ & $4.17_{\pm 4.03}$ \\
L-PPGN.L & $38.4_{\pm 6.4}$ & $32.6_{\pm 3.5}$ & $4_{\pm 3.6}$ & $3.9_{\pm 3.8}$ \\
GAG & \textbf{2.3$_{\pm 1.2}$} & $3.6_{\pm 1.3}$ & \textbf{0.5$_{\pm 1.5}$} & \textbf{0.1$_{\pm 1.7}$} \\
\bottomrule
\end{tabular}
  }
\end{table}

GAG generates text-rich dynamic graphs by collecting actor-item interaction data, simulating the process of distilling textual information into structured representations that resemble real-world networks.
To evaluate whether the generated graphs preserve the text-structure correlations of the seed graph, we adopt Graph Neural Networks (GNNs) benchmarking tasks of node classifications~\cite{yoon2023graph}. 
Specifically, we train different GNN architectures on both the generated graph and the original graph, measuring the accuracy gap between the two, denoted as $\Delta{Acc}$.  A smaller \( \Delta{Acc} \) indicates better preservation of the text-structure correlations.
For benchmarking, we select four representative GNN architectures and construct graphs with eight distinct relationships between graph structures and textual features. 
Details of the experimental setup are provided in Appendix \ref{section:ablation_text}.
As shown in Table~\ref{tab:performance_comparison_nc}, the random baselines perform the worst, 
highlighting the importance of modeling the tight coupling between graph textual feature and structure. 
Existing models such as GRAN and GraphMaker demonstrate strong performance in GNN benchmarking tasks, consistent with the findings in \cite{li2024graphmaker}. However, GAG consistently outperforms these baselines, achieving an average improvement of 1.45 in $\Delta{Acc}$ across GNNs, with $\Delta{Acc}$ values ranging from 0.09 to 3.61. 
These results demonstrate GAG's effectiveness in capturing intricate text-structure correlations.

\subsection{Scalability of GAG}
\label{section:gag_scale}

\begin{table}[htbp]
  \centering
  \caption{The time cost (\textit{min}) of 40 actor agents to interact once with the item agent.}
    \label{table:time_cost_port}
    \centering
    \resizebox{.6\linewidth}{!}{%
    \begin{tabular}{rlll}
      \toprule
      $P$ & SC & TC & SoC \\
      \midrule
      1 & 3.6250 & 0.0683 & 0.0623 \\
      4 & 0.1470 & 0.0068 & 0.0112 \\
      16 & 0.1160 & 0.0053 & 0.0109 \\
      24 & 0.0910 & 0.0054 & 0.0060 \\
      \midrule
      1$\rightarrow$24 & $\downarrow$\textbf{97.5\%} 
      & $\downarrow$\textbf{92.1\%} & $\downarrow$\textbf{90.4\%} \\
      \bottomrule
    \end{tabular}
    }
\end{table}
  
  We assess GAG scalability in terms of both graph generation scale and efficiency.
  Regarding generation scale, as shown in Table~\ref{table:clustering}, GAG supports large-scale graph generation of up to nearly 100,000 nodes in action network, or 12.2 million edges in 
  bib-coupling network. 
  In contrast, existing graph generation models're limited to 5,000 nodes~\citep{bergmeister2024efficient, efficient} or sparse grid graphs~\citep{dai2020scalable}, the detailed comparison with existing graph generation models is presented in Appendix \ref{appendix:section_scalibity}.
  Regarding generation efficiency,
we evaluate the effectiveness of parallel acceleration by examining the impact of \( P \) on the runtime performance of the GAG framework. Experiments are conducted on a machine equipped with 96 CPU cores and 376GB of memory.  
As shown in Table \ref{table:time_cost_port}, when the number of actor agents (\( n \)) is held constant, the time required to generate one item-actor interaction data is reduced by at least 97.5\% when \( P > 1 \) compared to \( P = 1 \). This highlights the effectiveness of parallel acceleration, and the capability of GAG in supporting simulation of large-scale graph evolution.~\footnote{
  We visualize the graph evolution process in \url{https://anonymous.4open.science/r/GraphAgent-2206/visualization/social_network.mp4}}

\section{Conclusion}
In this study, we present GAG, a novel and general framework designed for generating dynamic large-scale text-rich graphs with human interaction simulation. The generated graphs exhibit seven macro-level characteristics of real-world networks, including power law, small world and shrinking diameter.
In the graph expansion task, GAG surpasses existing baselines in graph expansion tasks by 11\% on specific evaluation metrics.
Furthermore, we present the S-RAG algorithm for simulating diverse human interaction processes at scale, complemented by parallel acceleration for simulation speed-up, achieving a speed-up of at least 90.4\%. Our framework successfully produces high-quality graphs with up to nearly 100,000 nodes or 10 million edges. 
Overall, GAG represents a promising initial step toward the efficient generation of dynamic, large-scale, text-rich graphs.

\section*{Limitations}
This paper acknowledges several limitations that future research could address:

\textbf{Behavior Interpretability} 
Though previous work has demonstrated that persona-enhanced prompting effectively guides LLMs in generating distinctive role-play synthetic data~\citep{chan2024scalingsyntheticdatacreation}, the mechanism of in-context learning remains a black box. Specifically, it remains unclear which prompts instruct agents to exhibit heterogeneous behavior, a challenge in LLM-based simulation works. Existing approaches explore the prompt-behavior correlation in LLMs through techniques like Knowledge Circuit~\cite{yao2024knowledge} and SAE-based representation engineering~\cite{zhao2024steering}. In the future, we aim to integrate such methods to provide layer-level explanations of LLM parameters during the simulation process, helping to explain the diverse behaviors of LLM-based agents during actor-item interactions.

\textbf{Simulation Scenario}
We acknowledge that simulation-based graph generation is currently suitable only for social networks. For domains such as point clouds, traffic networks, and molecular graphs, GAG is not directly applicable. However, as LLMs store a significant amount of factual knowledge in their parameters, we believe that with the continued development of LLM capabilities, the simulation of LLM-based agents can be extended beyond human behavior to other domains. For example, LLM-based agents could simulate molecular functional groups, interacting agent-wise to form chemical bonds, thereby generating molecular graphs. Based on this, GAG could be expanded to a broader range of applications.

% \section*{Ethics Statement}
% This work fully complies with the ACL Ethics Policy. To the best of our knowledge, we declare that there are no ethical issues in this paper.

\bibliography{custom}
\clearpage

\appendix

\section{Details of GAG}

\begin{table*}[thbp]
  \centering
  \caption{The type of actors and items for different simulation scenarios.}
  \label{table:simulation_setting}
  \resizebox{\textwidth}{!}{%
  \begin{tabular}{l|llll}
    \toprule
    % & \multicolumn{2}{c}{Agent}  &  $C$\\
    Scenario & Seed Graph & Actor Type & Item Type & Action Type\\
    \midrule

    \multirow{2}{*}{SC} & Citeseer~\citep{CiteGraph_Sen_08},  & Paper Author & Papers & Creation, Citation\\
    &Cora~\citep{CiteGraph_Sen_08}& & &\\
    &LLM-Generated& & &\\
    \midrule
    \multirow{2}{*}{TC} & Movielens~\citep{Harper2016TheMD}, & Movie Watcher & Movies & Rating\\
    &LLM-Generated& & &\\
    \midrule
    % TC & {Movielens~\citep{harper2015movielens},LLM-Generated} & Movie Watcher & Movies & Movie Rating\\
    SoC & {LLM-Generated} & Twitter User & Tweets & Tweet, Retweet, Reply, Follow\\
    \bottomrule
    \end{tabular}
    }
  
\end{table*}

To demonstrate the versatility of the GAG Framework, we build three graph generation tasks for different human activities in our experiments, the concrete settings are as follows:
(1) SC: In this scenario, llm-based agents act as authors interacting with a paper database and generate the following networks: paper citation, bibliographic coupling (bib-coupling), co-citation, author citation, and co-authorship networks~\citep{garfield2000web}.
(2) TC: In this scenario, llm-based agents act as reviewers interacting with a movie database and generate the following networks: the movie rating and the user projection networks~\citep{zhou2007bipartite}.
(3) SoC: In this scenario, llm-based agents act as users interacting with a twitter-like social media database and generate the following networks: follow, friend, and action networks~\citep{de2013anatomy}. 

\subsection{Node Formulation}

We initilize the original item and actor node set from $B_0$. We use Citeseer as $B_0$ for SC and Movielens as $B_0$ for TC. 
For SoC, we use LLM-generated graph textual features to construct the initial item and actor node set, eliminating the need for external data collection. Further details regarding the configuration of the actors and items are summarized in Table \ref{table:simulation_setting}.

\paragraph{Item Node} 
\label{appendix:item_node}
Since the seed graph \( B_0 \) lacks certain text-rich item nodes and actor nodes (e.g., the Citeseer dataset is missing author information and article content), we crawl the necessary node attributes to enrich the text attributes for Citeseer~\cite{CiteGraph_Sen_08}. The text-enriched dataset is available in the open-source repository.

For the \( k \)-th (\( k > 1 \)) simulation round, we employ an item template that maps each item node \( v_{j,k-1} \) to the corresponding text attribute vector \( x_{j,k-1}^V \)~\cite{taglas}. This graph-to-text transformation process leverages the node features and optionally considers edge features to enhance the representation.
For instance, in a citation network, graph nodes represent academic papers, whereas in a movie-rating network, they represent movies.  
The item templates for each simulation scenario are detailed below:
\begin{table}[H]
  \caption{The item template of papers in SC.}
  \label{prompt:paper}
  \begin{tcolorbox}[colback=blue!5, % Light blue background
  colframe=black]
  \textbf{Node Feature}: Academic paper. 

  Title: $<$Title$>$

  Topic: $<$Topic$>$
  
  Abstract: $<$Abstract$>$

  \textbf{Edge Feature}: The citation/writing relationship connecting papers and authors.
  \end{tcolorbox}
\end{table}

\begin{table}[H]
      \caption{The item template of movies in TC.}
      \label{prompt:movie}
      \begin{tcolorbox}[colback=blue!5, 
      colframe=black,
      ]
      \textbf{Node Feature ($v_i, x_i$)}: Movie. 
      
      Title: $<$Title$>$ 
      
      Genres: $<$Genres$>$ 
      
      Content: $<$Movie Abstract$>$ 

      \textbf{Edge Feature}: The movie rating data connecting watchers and movies.
        
      % \textbf{Edge Feature ($E_i$)}: NULL.
    % \textbf{Edge Feature ($E_i$)}: The movie ratings connecting the movie watchers and the movie $v_i$.
      \end{tcolorbox}
\end{table}

 \begin{table}[H]
      \caption{The item template of tweets in SoC.}
      \label{prompt:tweet}
      \begin{tcolorbox}[colback=blue!5, 
      colframe=black,
      ]
    \textbf{Node Feature ($v_i, x_i$)}: Tweets. 
    
    Tweet ID: $<$Tweet ID$>$ 
    
    User: $<$Tweet User$>$ 
    
    Tweet: $<$Tweet Content$>$ 

    \textbf{Edge Feature}: The tweet history connecting tweets and tweet users.
    \end{tcolorbox}
    \end{table}

\paragraph{Actor Node}
\label{appendix:agent_formulation}

In the \( k \)-th simulation round, we generate a specified number of role-playing synthetic profiles, referred to as \( \widetilde{X_{k}^A} \). 
For the various simulation scenarios, we develop LLM-based agents to form the actor node set, each assigned distinct roles such as paper authors, movie watchers, or Twitter users. These agents engage with an item set through a predefined set of actions. 

Each round includes the addition of profiles, with 30 actor profiles for the SC simulation and 25 for the SoC simulation. For the TC simulation, actor profiles are dynamically added based on node timestamp information from the Movielens. The prompts used for profile generation are outlined in Table~\ref{table:profile_prompt_sc}, Table~\ref{table:profile_prompt_tc}, Table~\ref{table:profile_prompt_soc}.

To determine the action state of each agent, we employ two distinct strategies. For the SC simulation, the number of active agents is proportional to the number of papers created in that round (set at 50). For the TC simulation, all actors remain active. 
For the SoC simulation, we use a different approach. Since in online social media networks, the influence of content shared can vary significantly between core users (those with a higher level of engagement or influence) and general users. Research indicates that core users typically make up around 20\% of the entire user base in a social graph, following the Pareto distribution principle~\citep{mislove2007measurement, mou2024unveiling}. We categorize the core users as agents labeled as \textit{core}, denoted as \( HUB \). This characterization allows us to analyze the dynamics of influence within simulated environments. We further explore variations by adjusting the ratio of LLM-based agents designated as \textit{core} in Appendix \ref{appendix:coarse_ranking}.

\subsection{Interaction Simulation}
\label{appendix:pairwise_interaction}
% 为了确定与a_l最相关的条目信息,我们提出了S-RAG algorithm. 完整的算法如Algorithm \ref{alg:s-rag_complete}所示。

\paragraph{Interaction Process}
In the $k$-th simulation round, to identify the most relevant information for \( a_{i,k} \), we propose the S-RAG algorithm. As actor is prompted to give the intial query set $Q_{i,k}$, we collect the feedback $O_{i,k,q}$ for every query $q \in Q_{i,k}$.
These feedbacks collectively form the environment feedback set $O_{i,k}$.
For detailed explanation of the query process for one query to obtain $O_{i,k,q}$ in S-RAG: 

1. In the recall stage, we initially retrieve \( O_{i,k,q} \) as the candidate documents, which serves as the initial environmental feedback. This step we filter out $N_r$ candidate documents.

2. In order to align $O_{i,k,q}$ with agent's personal preference more accurately, we adopt the reranking stage for post-processing of $O_{i,k,q}$, which is divided into two phases:
(1) Coarse Ranking: We reorder \( O_{i,k,q} \) based on whether the interaction data was generated by agents labeled as \textit{core}. Candidate documents originating from \textit{core} agents are positioned at the forefront of \( O_{i,k,q} \), while those from non-core agents are placed towards the end.
(2) Fine Ranking: 
We further reorganize \( O_{i,k,q} \) based on the personal preferences of the agents. For SC simulation, the filter items include topics of the academic papers; for TC simulation, the filter items include movie genres; and for SoC simulation, the filter items include attributes of posted tweets, such as friends, topics, and follows. Ablation study on the filter items is detailed in Appendix. \ref{appendix:ablation_filter}.

3. In the generation stage, we prompted actor nodes to act accordingly based on their observations. To this end, we define different action prompt templates based on the simulation scenario. The action prompt templates are defined in Table \ref{prompt:paper_writing}, Table \ref{prompt:movie_rating}, and Table \ref{prompt:tweet_sending}.

\paragraph{Parallel Acceleration}
\label{Appendix:nested_actor}
To enhance the simulation speed of GAG, we propose Nested-ACTOR based on the traditional actor architecture~\citep{actor}. As highlighted by \citep{clauset2004finding}, network structures often exhibit densely connected communities with weaker inter-community links. To exploit this characteristic, the primary goal is to categorize agents into groups, each defined by an active actor agent paired with an item agent, thereby enabling parallel execution across these groups. We initialize a supervisor agent to manage the agents within each group, with each supervisor actor assigned to a different CPU core of the computational machine. Between groups, the item agent and active actor agent share a single message queue to facilitate intra-group message processing. Within groups, supervisor actors manage inter-group parallel message processing.
Agents within a single group only need to account for the I/O wait times of other agents in that group, rather than waiting on all agents in the system.

\subsection{Graph Projection}
\label{Appendix:graph_projection}

In various simulation scenarios, 
$B_0$ progressively evolves into $B_K$ after $K$ rounds of interaction simulations.
As defined in Table~\ref{table:simulation_setting}, the bipartite graph are marked by nodes and edges of different types. Specifically, the action type marks the edge type; the item and actor type marks the node type; and the associated textual attribute marks the node attribute. The sub-graph, denoted as $G(\mathcal{V}^s,\mathcal{E}^s)$, is projected by $B_{K}$. Following established folding rules in network science research, the sub-graphs embody different semantic interpretations. 

In the context of SC, following action template in Table \ref{prompt:paper_writing}, each time the author generates a paper and references other papers. To fold graphs from the pair-wise interaction process, we define the following mapping functions:

\begin{enumerate}
    \item \textbf{Paper Citation:} Let \( \mathcal{V}^s\) represents the set of papers, \( \mathcal{E}^s\) represents the one paper is cited by another paper of one author, and \( \mathcal{X}^{Vs}\) signify the textual attributes of each paper.
    
    \item \textbf{Bib Coupling:} Let \( \mathcal{V}^s \) represents the set of papers, \( \mathcal{E}^s\) represents the relationships where two papers cite the same reference, and \( \mathcal{X}^{Vs} \) encompasses the attributes of the papers.
    
    \item \textbf{Co-citation:} Let \( \mathcal{V}^s \) represents the set of papers, \(\mathcal{E}^s \) represents the relationships where two papers are cited by the same paper, and \( \mathcal{X}^{Vs} \) includes the attributes of the respective papers.
    
    \item \textbf{Author Citation:}  Let \(\mathcal{V}^s\) represents the set of authors,  \( \mathcal{E}^s \) represents the papers of one author is cited by another author, and \( \mathcal{X}^{Vs} \) refers to the attributes of each author.
    
    \item \textbf{Co-Authorship:} Let \( \mathcal{V}^s \) represents the set of authors, \( \mathcal{E}^s \) represents the collaborative relationships between authors, and \( \mathcal{X}^{Vs} \) characterizes the attributes of each author.
\end{enumerate}

In the context of TC, following action template in Table \ref{prompt:movie_rating}, each time the user generates a movie rating. To fold graphs from the pair-wise interaction process, we define the following mapping functions:

\begin{enumerate}
    \item \textbf{Movie Rating}: Let $\mathcal{V}^s$ represents the movie watchers and the movies, $\mathcal{E}^s$ represents the movie ratings. For movie watchers, $\mathcal{X}^{Vs}$ correspond to the attributes of movie watchers; for movies, $\mathcal{X}^{Vs}$ correspond to the attributes of movies.
    \item \textbf{User Projection}: Let $\mathcal{V}^s$ represent the movie watchers, $\mathcal{E}^s$ represents the movie watcher relationships who jointly rated movies, and $\mathcal{X}^{Vs}$ encompasses the attributes of the movie watchers.
\end{enumerate}

In the context of SoC, following action template in Table \ref{prompt:tweet_sending}, each time the user generates a tweet and retweet/reply other tweets and follow other users. To fold graphs from the pair-wise interaction process, we define the following mapping functions:

\begin{enumerate}
  \item \textbf{Action:} Let $\mathcal{V}^s$ represent users, $\mathcal{E}^s$ denote the edges indicating tweets exchanged between two users (e.g., retweets, follow, reply actions), and $\mathcal{X}^{Vs}$ represent user attributes.
  \item \textbf{Follow:} Let $\mathcal{V}^s$ represent users, $\mathcal{E}^s$ denote the edges indicating a follow relationship between two users, and $\mathcal{X}^{Vs}$ represent user attributes.
  \item \textbf{Friend:} Let $\mathcal{V}^s$ represent users, $\mathcal{E}^s$ denote the edges indicating a friend relationship between two users (i.e., mutual following), and $\mathcal{X}^{Vs}$ represent user attributes.
\end{enumerate}

\section{Graph Structure Alignment}

For the LLM backbone, we have chosen the open-source model of Llama-3-70B\cite{llama3modelcard}  for the large-scale graph generation in macro-level structure alignment experiment. 
For micro-level structure alignment, we select the closed-source model of GPT-3.5-turbo for a more accurate simulation of human behaviors. Additionally, we select \cite{reimers-2019-sentence-bert} \footnote{https://huggingface.co/sentence-transformers/all-MiniLM-L6-v2} as the $\operatorname{encoder}$ in S-RAG.

\subsection{Graph Structure Metrics}
\label{appendix:evaluation_metrics}

To measure the structural characteristics of graph,
we use the following structural metrics:

\textbf{(1) \(|\mathcal{V}^s|\)}: measures the node number of graph $\mathcal{G}^s$.

\textbf{(2) \(|\mathcal{E}^s|\)}: measures the edge number of graph $\mathcal{G}^s$.

\textbf{(3) \(\bar{cc}\)}: average clustering coefficient, quantifies the degree to which nodes in a graph tend to cluster together \footnote{https://en.wikipedia.org/wiki/Clustering\_coefficient.}.

\textbf{(4) $r$}: assortativity, measures the similarity of connections in the graph concerning the node degree. \footnote{https://en.wikipedia.org/wiki/Assortativity}

\textbf{(5) $D_{e}$}: effective diameter, defined as the minimum number of hops in which a certain percentage (typically 90\% or 95\%) of all connected node pairs can be reached.

\begin{table*}[htbp]
  \centering
  % \caption{The structural metrics of the generated networks by GAG.}
  \caption{The structural metrics for graphs generated by GAG.}
  \label{table:graph_structure_metrics}
  \resizebox{.85\textwidth}{!}{%
  \begin{tabular}{llllll}
  \toprule
  & Citation & Bib-Coupling & Co-Citation & Author Citation & Co-Authorship \\
  \midrule
  \( |\mathcal{V}^s| \)              & 1.14e+04 & 1.09e+04 & 3.93e+03 & 5.01e+03 & 5.01e+03 \\
  \( |\mathcal{E}^s| \)              & 3.63e+04 & 1.22e+07 & 3.27e+04 & 2.41e+05 & 2.08e+04 \\
  \( \bar{cc} \)                    & 0.08 & 0.77 & 0.59 & 0.38 & 0.20 \\
  \( r \)                          & -0.10 & 0.09 & -0.10 & -0.18 & 0.32 \\
  \( D_{e} \)                    & 5.19 & 2.94 & 3.89 & 3.44 & 5.77 \\
  \midrule
  &Action & Follow & Friend & Movie Rating & User Projection \\
  \midrule
  \( |\mathcal{V}^s| \)              & 9.97e+04 & 9.96e+04 & 9.96e+04 & 4.17e+03 & 3.91e+03 \\
  \( |\mathcal{E}^s| \)              & 9.07e+05 & 1.53e+06 & 5.01e+05 & 3.25e+04 & 9.04e+05 \\
  \( \bar{cc} \)                    & 0.07 & 0.61 & 1.00 & 0.00 & 0.34 \\
  \( r \)                          & -0.03 & 0.06 & 0.59 & -0.54 & -0.11 \\
  \( D_{e} \)                    & 2.79 & 2.85 & 11.66 & 2.98 & 1.17 \\
  \bottomrule
  \end{tabular}
  }
  \end{table*}

  As shown in Table \ref{table:graph_structure_metrics}, we calculate the structural metrics of all generated networks. Similar to the assortative-mixing patterns discovered in real-world networks~\citep{PhysRevLett.89.208701}, the citation network exhibits negative assortativity, whereas the co-authorship network displays positive assortativity. Moreover, the generated networks exhibit a small diameter, \( D_e \in [1.17, 11.66] \), consistent with the \textit{six degrees of separation} phenomenon observed in real-world networks~\citep{shrinking_diameter, BRODER2000309}.
  The high $\bar{cc}$, combined with small $D_{e}$, confirms these networks exhibit small-world characteristics.

\subsection{Macro-Level Evaluation}
\label{Appendix:macro_level_evaluation}

\begin{figure*}[htbp]
  \centering
  \subfloat[Movielens Rating Network]   % 第一张子图的下标（注意：注释要写在[]中括号内）
    {
        \centering
        \label{fig:movielens_p}
        \includegraphics[width=0.5\linewidth]{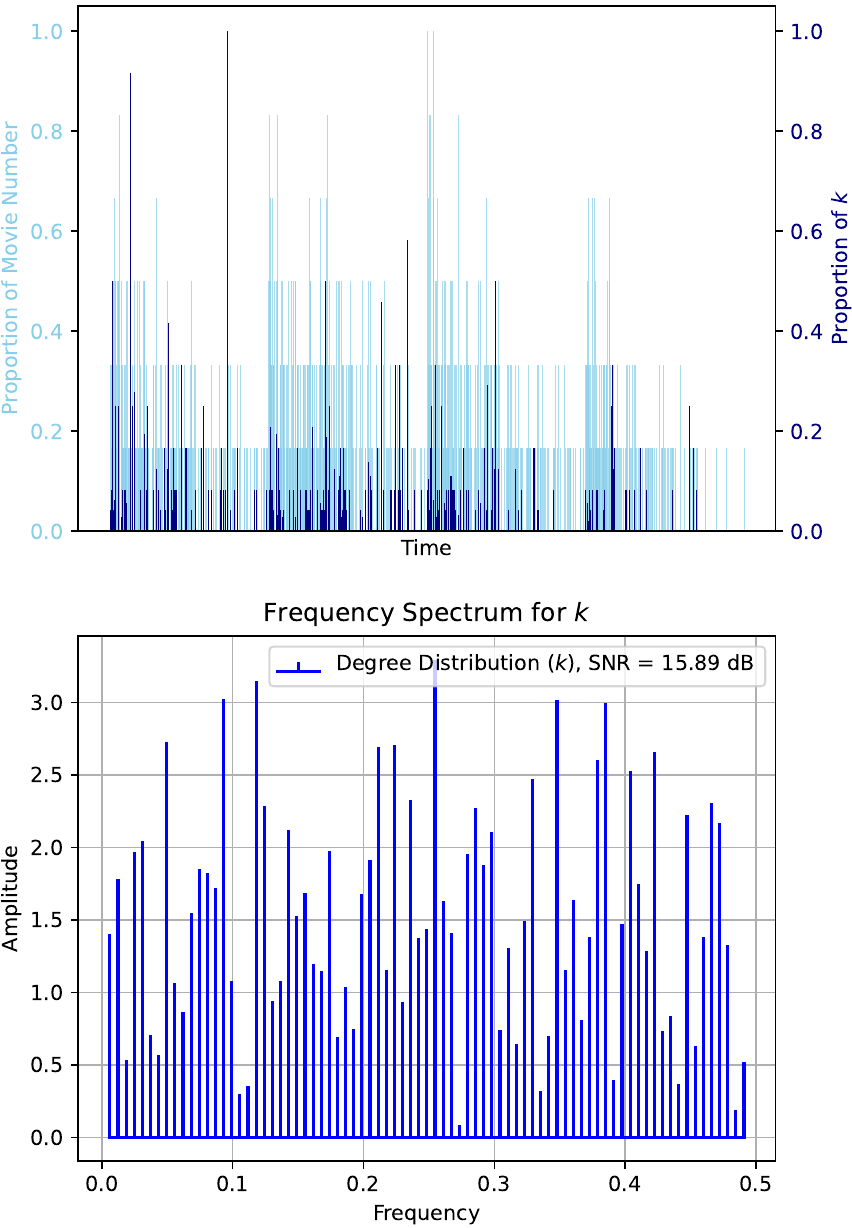}
    }
  \subfloat[Generated Movie Rating Network]   % 第一张子图的下标（注意：注释要写在[]中括号内）
    {
          \centering
        \label{fig:movie_rating_p}
        \includegraphics[width=0.5\linewidth]{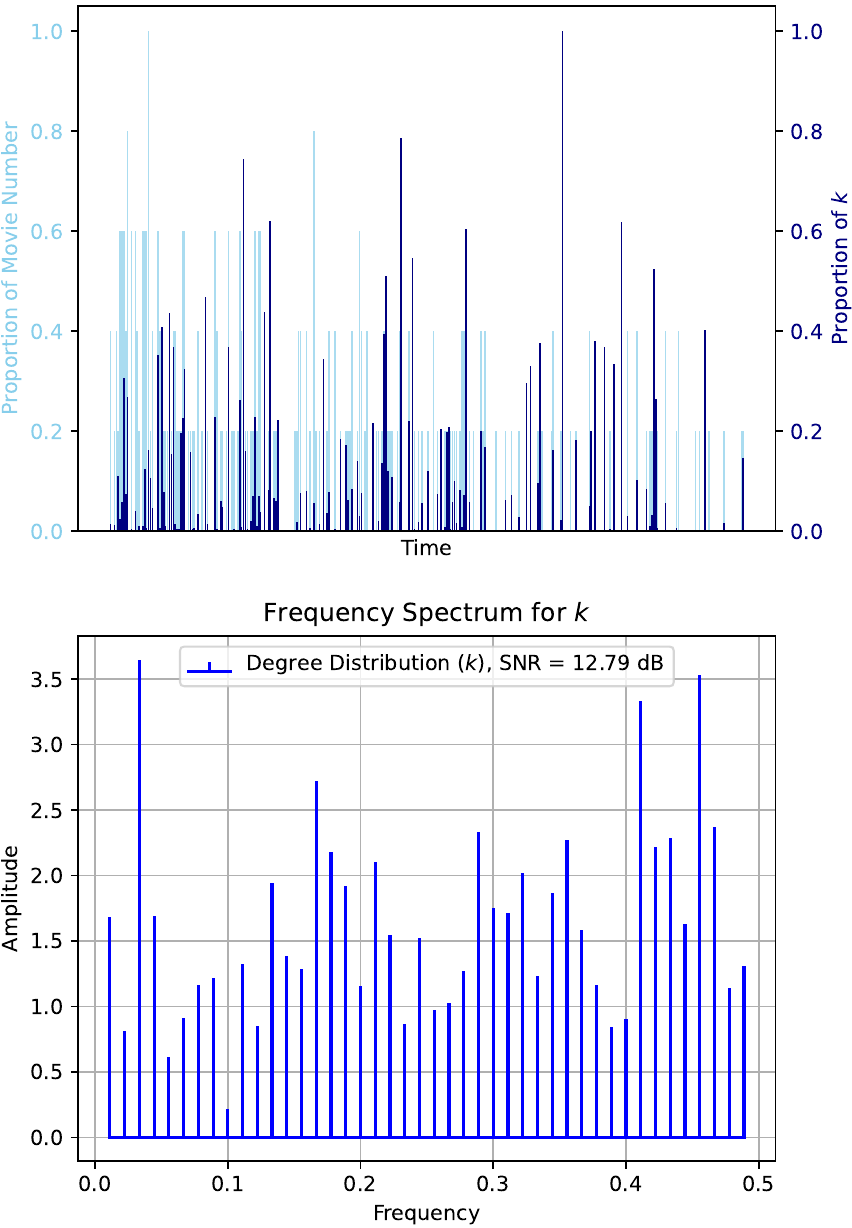}
    }
    
  \caption{Periodic Variation of Degree in Movie Rating Network; Figure \ref{fig:movielens_p} shows the number of released movies over time and the degree of the movie rating network over time in MovieLens dataset; Figure \ref{fig:movie_rating_p} also shows the number of released movies and the degree of the movie rating network over time in GAG.}
  \label{fig:periodic_flunc}
\end{figure*}

\paragraph{Periodic Variation of Degree}
In the simulation scenario of TC, we filter the review data from the Movielens-25M dataset based on the movies listed in the Movielens-1M dataset. We select the top 10 ratings for each user and discover a noteworthy phenomenon: the number of reviews in the rating network exhibits periodic fluctuations over time. By scraping the release dates of the movies and plotting their release frequency, we observe that the periodicity in the release frequency is consistent with the fluctuations in the number of reviews. To quantify the periodicity, we selected the signal-to-noise ratio (SNR)~\citep{johnson2006signal} as our metric, considering an SNR greater than 10 dB to indicate strong periodicity and reliability. Furthermore, we observe that the degree of the generated rating graph also exhibits periodic variations consistent with the release dates of the movies. As illustrated in Figure \ref{fig:periodic_flunc}, the SNR of the degree of the rating graph is 12.79 dB, surpassing the 10 dB threshold, thus demonstrating significant periodic fluctuations. 

\begin{figure}[htbp]
  \centering
  \includegraphics[width=\linewidth]{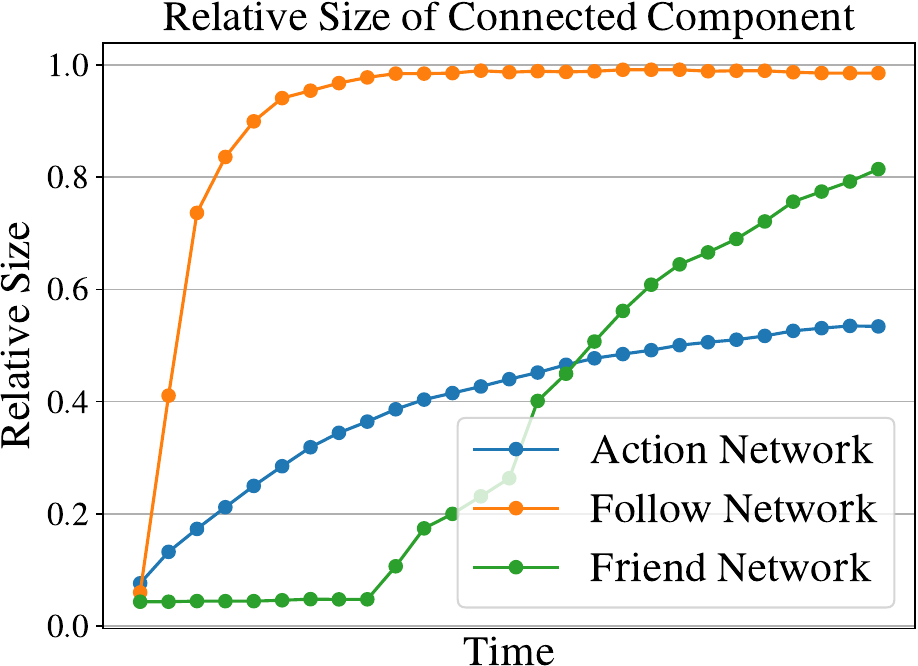}
  \caption{The proportion of the largest connected component grows steadily over time.}
  \label{fig:gcc_emergent}
\end{figure}

\paragraph{Emergent of GCC}
In online social graphs, nodes with higher degrees grow larger over time and eventually manifest a giant connected component (GCC)~\citep{mislove2007measurement}. As illustrated in Figure \ref{fig:gcc_emergent}, the proportion of the largest connected component rows steadily over time, indicating the emergence of a giant community within the social graph. The network is generated by with 7000 actor agents.
% In the friend network, as shown in Figure \ref{fig:shrinking_diameter}, the diameter exhibits sublinear growth and gradually decreases. This low $D_{e}$ further substantiates the small-world characteristics of the network. 

\paragraph{Friendship Paradox}
\begin{figure}[H]
  \centering
  \subfloat[Action Network]   % 第一张子图的下标（注意：注释要写在[]中括号内）
    {
          \centering
        \label{fig:action_friend}
        \includegraphics[width=\linewidth]{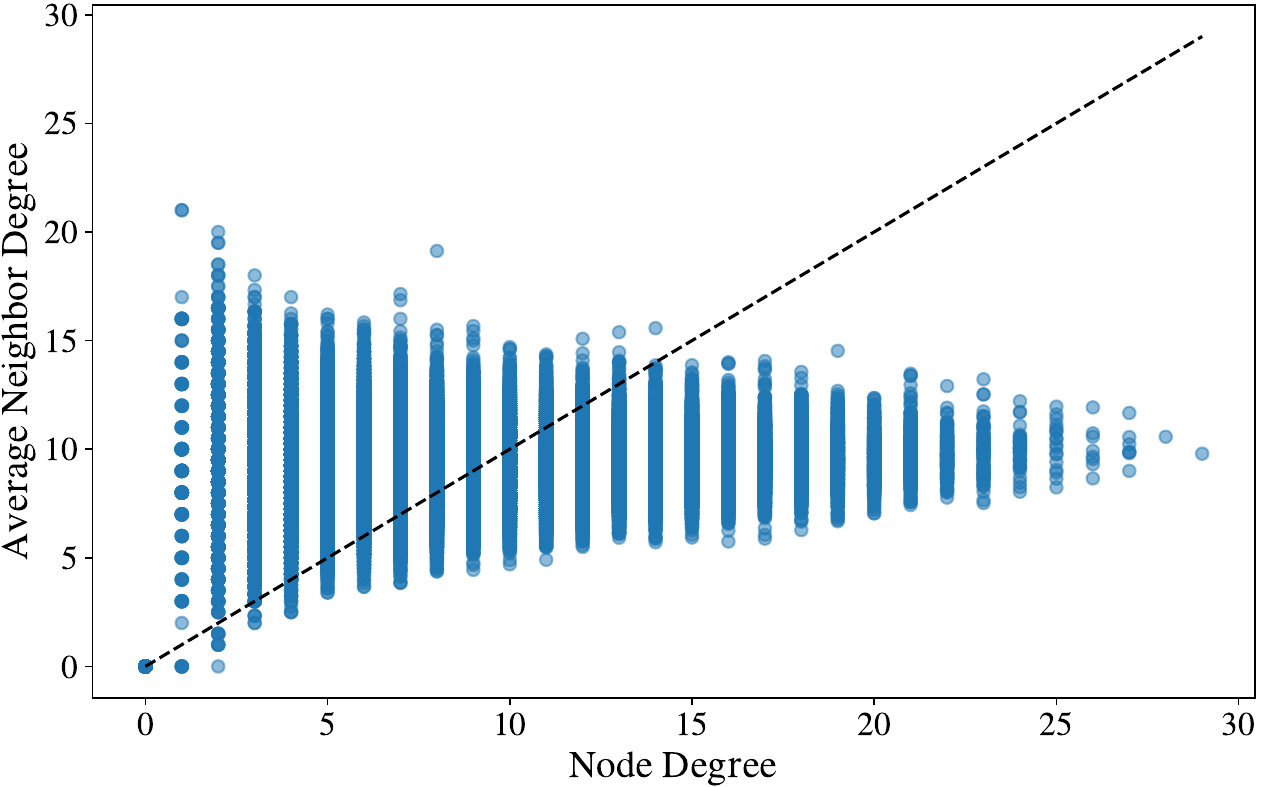}
    }
    \\
  \subfloat[Follow Network]    % 第一张子图的下标（注意：注释要写在[]中括号内）
    {
        \centering
        \label{fig:follow_friend}
        \includegraphics[width=\linewidth]{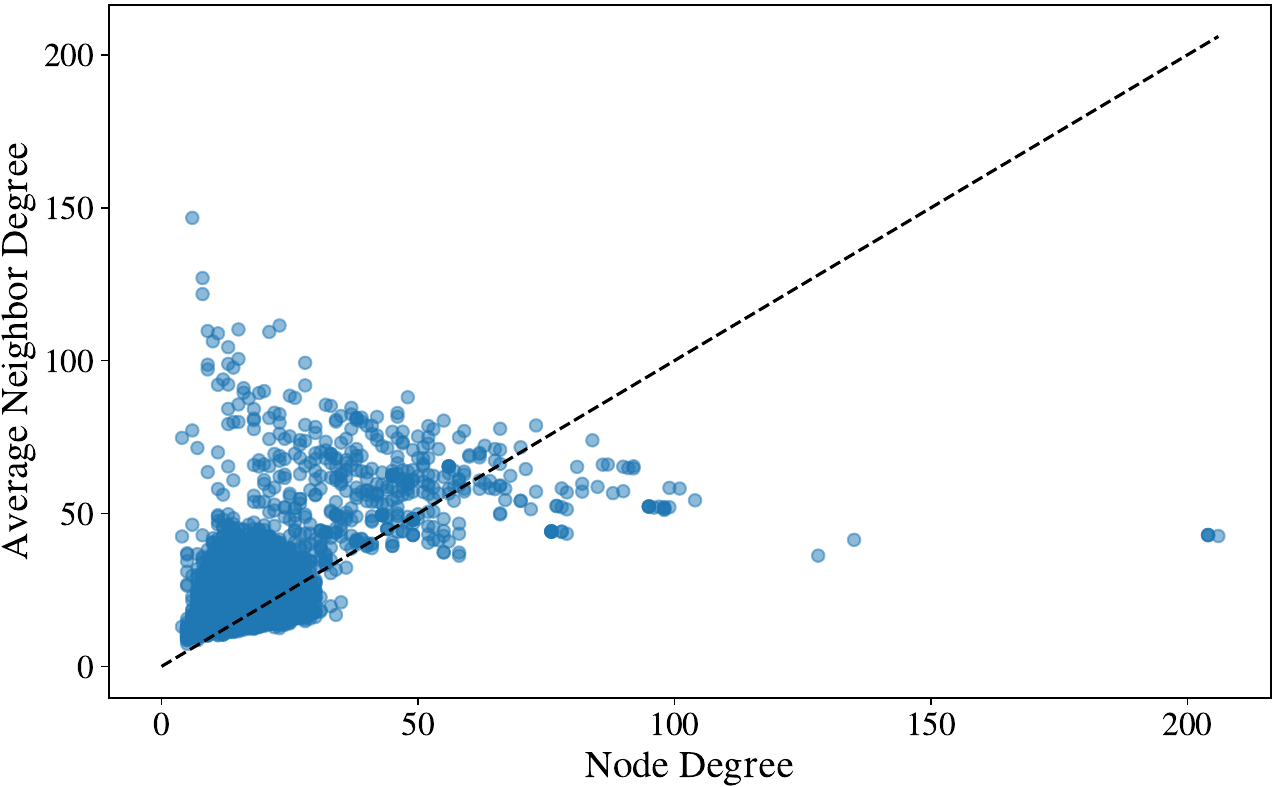}
    }
    \\
     \subfloat[Friend Network]    % 第一张子图的下标（注意：注释要写在[]中括号内）
    {
        \centering
        \label{fig:friend_friend}
        \includegraphics[width=\linewidth]{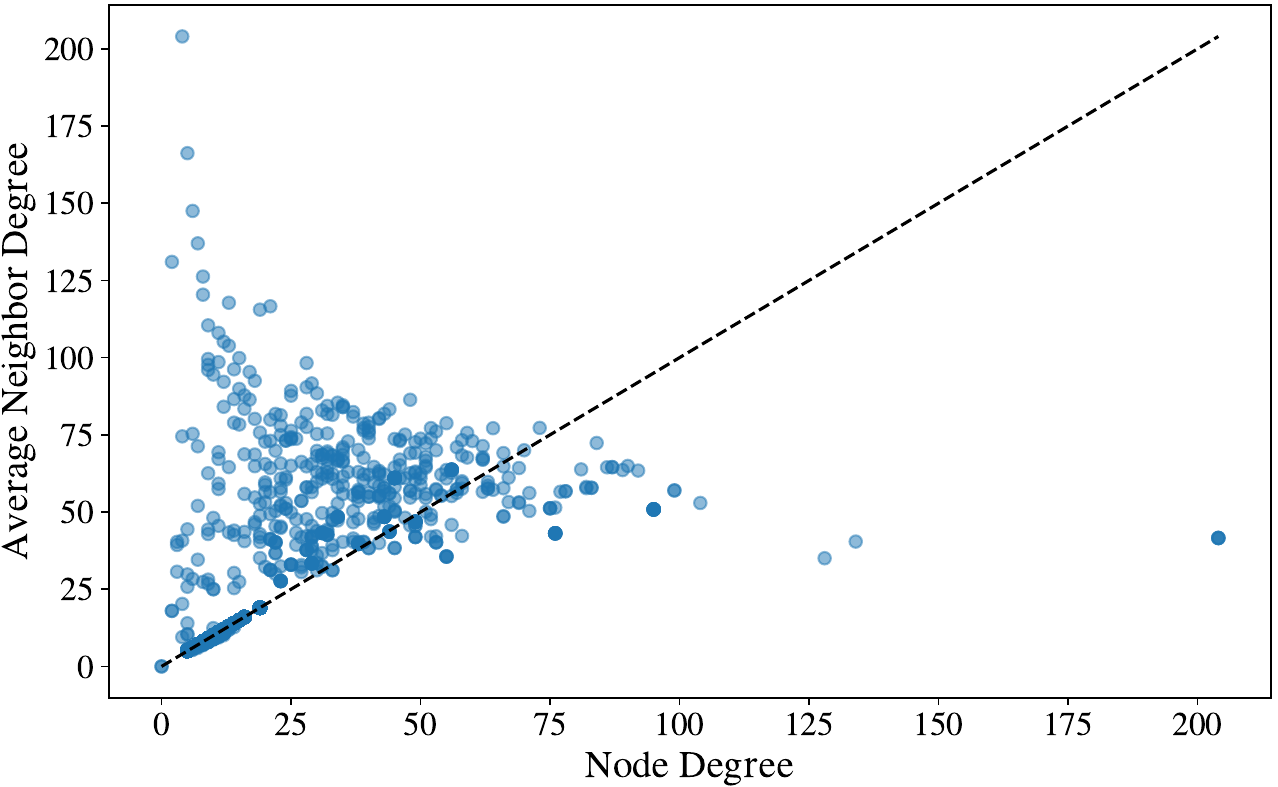}
    }
  \caption{The \textit{Friendship Paradox} phenomenon in online social graphs; The figures show the average degree of node neighbors v.s. the average degree of node itself in social graphs.}
  \label{fig:friendship_paradox}
\end{figure}

An interesting and somewhat counterintuitive phenomenon in real-world social graphs is that everyone you follow or who follows you tends to have more friends and followers than you do. This phenomenon has been observed in both Twitter~\citep{hodas2013friendshipparadoxreduxfriends} and the social graph of Facebook~\citep{ugander2011anatomy}, applying to more than 98\% of the nodes. As shown in Figure \ref{fig:friendship_paradox}, the friendship paradox is most evident in the friend network, with over 90\% of the nodes lying above the $y = x$ line, indicating that most users have fewer friends than their friends do. The network is generated by 5 rounds of simulation with 1e5 agents.

\paragraph{Densely Connected Core}

\begin{figure}[htbp]
  \centering
  \includegraphics[width=\linewidth]{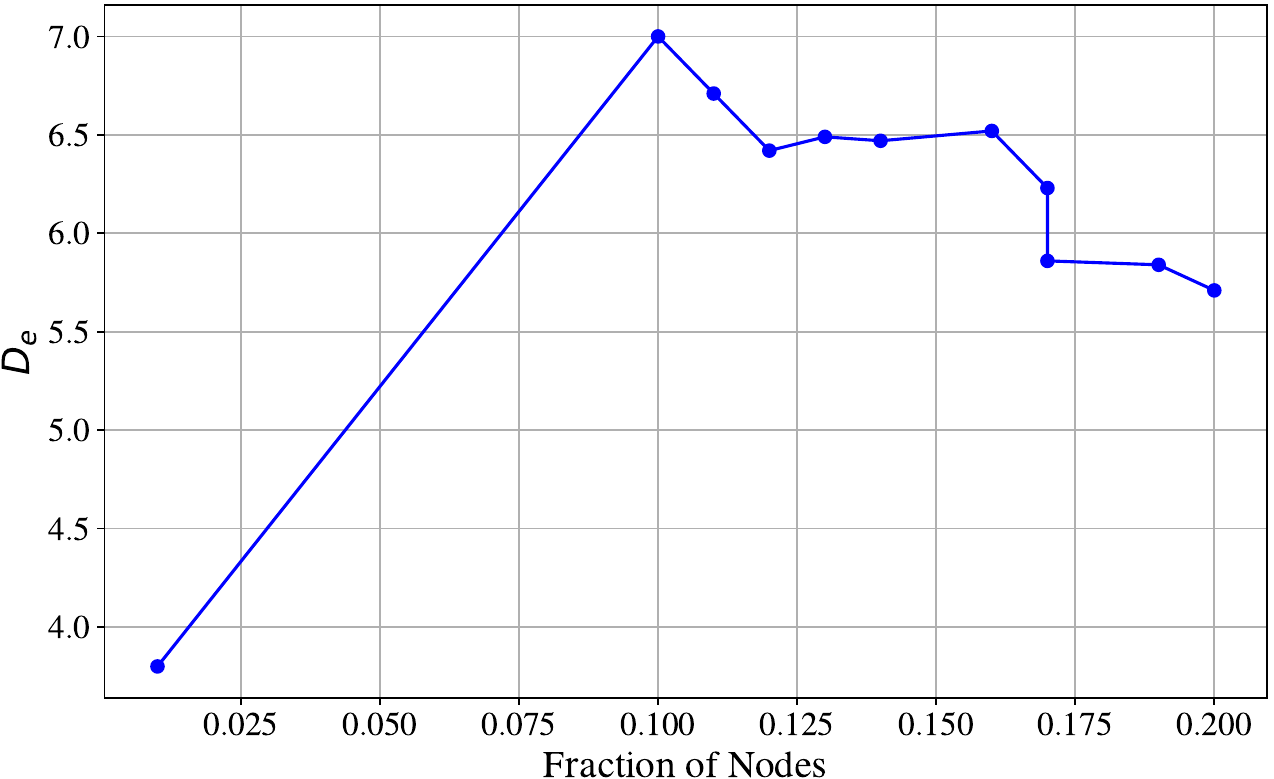}
  \caption{The out-degree is plotted against $\bar{cc}$ in follow graph.}
  \label{fig:friend_avg_path}
\end{figure}

\begin{figure}[htbp]
  \centering
  
  \includegraphics[width=\linewidth]{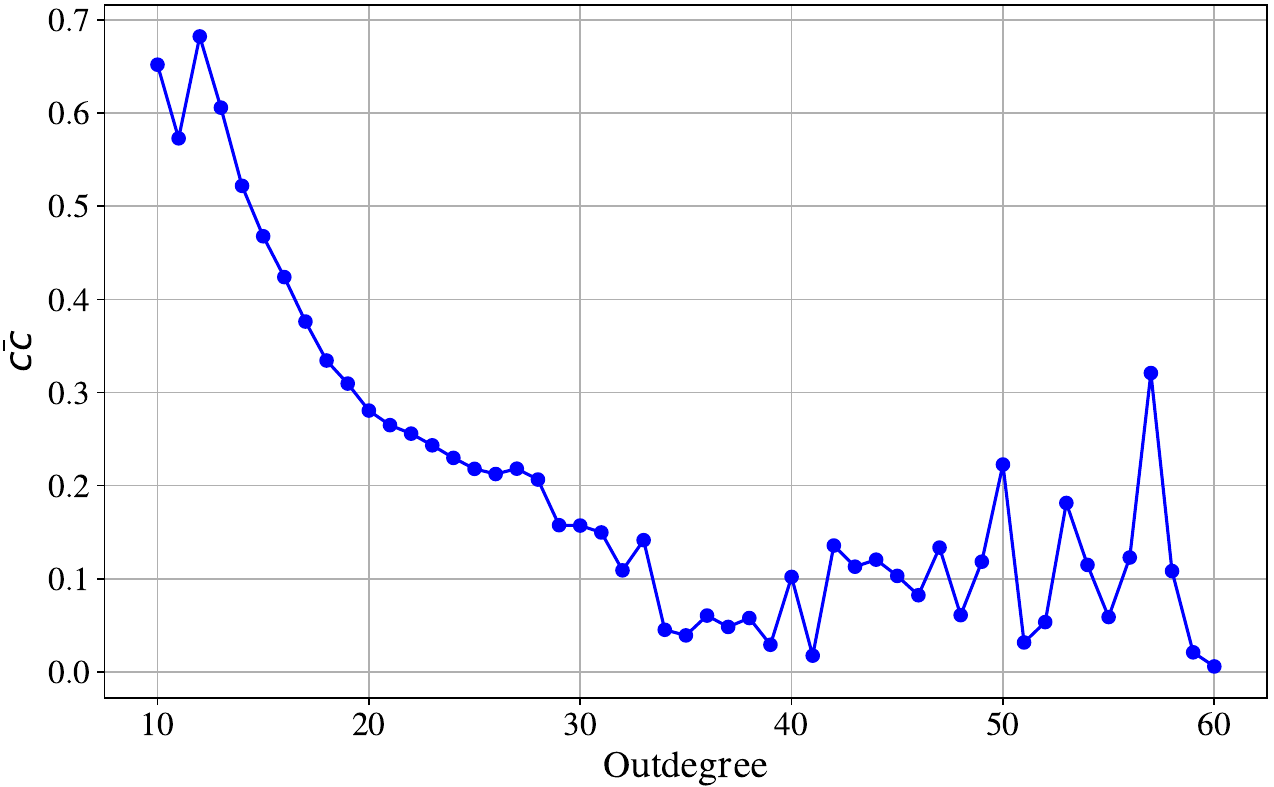}
  \caption{The $D_e$ of the DCC in follow network, the network is divided by the fractions of the total number of nodes.}
  \label{fig:degree_cc}
\end{figure}

In real-world online social graphs, there exists a densely connected core (DCC) comprising between 1\% and 10\% of the highest degree nodes, such that removing this core completely disconnects the graph~\citep{mislove2007measurement}. These high-degree nodes serve as hubs of the network, causing the network to become increasingly compact through the hub structure. As shown in Figure \ref{fig:degree_cc}, the nodes with higher degrees have significantly higher $\bar{cc}$ compared to other nodes. For these densely connected components, $D_e$ grows at a slow rate.
In Figure \ref{fig:friend_avg_path}, we observe that the $D_e$ among the DCC of follow network is grows sublognitively. The network is generated by 5 rounds of simulation with 1e5 agents.

\subsection{Micro-Level Evaluation}
\label{appendix:micro_level_evaluation}

GAG distinguishes itself from traditional graph generation methods by generating graph data without requiring prior training. It achieves this through the simulation of human behavior, leading to the emergence of various structural features characteristic of real-world networks. Consequently, conventional graph evaluation methods cannot be applied. To address this, we design comparative experiments against existing graph generation models. 

\begin{table*}[htbp]
  \centering
  \caption{Training hyperparameters of baseline models. All unspecified hyperparameters default to their standard values.}
  \label{table:hyper_params}
  \resizebox{\linewidth}{!}{
  \begin{tabular}{lll}
  \toprule
  Model & Hyperparameter & Experiment\\
  \midrule
  Erdös-Rényi~\citep{erdos1960evolution}
    & Linking propoblity & \( \bar{k} / (|\mathcal{V}^s| - 1) \) \\
  \midrule
  Barabási-Albert~\citep{barabasi1999emergence}
    & Number of linking edges & \( \bar{k} / 2 \) \\
  \midrule
  Small-World~\citep{watts1998collective}
    & Number of linking nodes & \( \bar{k} \times 2 \) \\
  \midrule
  BiGG~\citep{dai2020scalable}
    & Ordering & DFS \\
    & Accumulated gradients & 1\\
    & Batch size & 32 \\
    \midrule
  GRAN~\citep{efficient}
    & Hidden size & 512 \\
    & Embedding size & 512\\
    & Number of layers & 7 \\
    & Number of mixtures & 20 \\
    & Batch size & 16\\
  \midrule
  BwR~\citep{diamant2023improving}
    & Model & GraphRNN~\citep{you2018graphrnn}\\
    & bw & 8 \\
    & Hidden size & 128\\
    & Ordering & BFS \\
    & Batch size &32 \\
  \midrule
  L-PPGN~\citep{bergmeister2024efficient}
    & Hidden embedding size & 256 \\
    & PPGN embedding size & 128\\
    & Input embedding size & 32\\
    & Number of layers & 10 \\
    & Number of denoising steps & 1024\\
    & Batch size & 32\\
    & EMA coefficient & 0.99\\
    & Number of spectral features & 0 \\
  \midrule
  GraphMaker~\citep{li2024graphmaker}
    &Variant & Sync \\
    &Hidden size for timestep &32\\ 
    &Hidden size for node & 512\\
    &Hidden size for node label &64\\
    &NumberofMPNNlayers &2\\ 
    &Learning rate &0.001\\
    & Optimizer &AMSGrad\\
  
  \bottomrule
  \end{tabular}
  }
  \end{table*}

  \begin{table*}[h]
    \centering
    \caption{Comparison with existing expansion-based graph generation models. For GRAN, generated graph degree distribution fails to converge when fitting a power-law distribution.}
    \label{table:compare_structure_cora}
    \resizebox{\textwidth}{!}{%
      \begin{tabular}{lrrrrrrrr}
      \toprule
      &  MMD.D$\downarrow$ &  MMD.C$\downarrow$ &  MMD.S$\downarrow$ &  MMD.O$\downarrow$ &  $D_{k}$ & $\alpha$ &  Valid$\uparrow$ &GEM\\
      \midrule
      Cora        &    - &   - & - & -  &$0.07_{\pm 0.0}$ &  $2.59_{\pm 0.01}$ &   1.00 &   - \\
      \midrule
      Erdös-Rényi    &        0.25 &         1.41 &         0.54 &       0.27 &  $0.13_{\pm 0.02}$ &  $4.01_{\pm 0.17}$ &   0.00 &  0.29 \\
      Barabási-Albert       &        0.09 &         1.41 &         0.44 &       1.11 &  $0.04_{\pm 0.01}$ &   $2.4_{\pm 0.05}$ &   1.00 &  0.46 \\
      Small-World  &        0.60 &         1.41 &         0.50 &       0.20 &   $0.14_{\pm 0.0}$ &  $4.05_{\pm 0.02}$ &   0.00 &  0.28 \\
      BiGG        &        0.14 &         0.51 &         0.48 &       0.27 &  $0.08_{\pm 0.01}$ &  $3.19_{\pm 0.11}$ &   0.05 &  0.34 \\
      GRAN          &        0.15 &         0.50 &         0.55 &       0.28 &    - &   - &   0.35 &  0.40 \\
      BwR      &        0.32 &         0.23 &         0.34 &       0.19 &   $0.1_{\pm 0.01}$ &  $3.58_{\pm 0.08}$ &   0.00 &  0.35 \\
      GraphMaker   &        0.37 &         1.41 &         0.75 &       0.28 &    - &    - &   0.00 &  0.27 \\
      L-PPGN     &        0.15 &         0.92 &         0.32 &       0.59 &  $0.06_{\pm 0.01}$ &  $2.77_{\pm 0.04}$ &   1.00 &  0.50 \\
      GAG     &        0.35 &         0.84 &         0.41 &       1.21 &   $0.09_{\pm 0.0}$ &   $2.1_{\pm 0.01}$ &   1.00 &  0.47 \\
      \bottomrule
      \end{tabular}
      }
  \end{table*}

\paragraph{Evaluation Metrics}
In accordance with established evaluation metrics for graph generation \cite{bergmeister2024efficient}, we report the maximum mean discrepancy (MMD) between the generated graphs and the test graphs, specifically focusing on degree distribution and clustering coefficient. Furthermore, we place particular emphasis on evaluating whether the degree distribution of the generated graphs conforms to a power law after expanding the graph to out-of-distribution sizes \cite{clauset2009power}. 
To this end, we employ six key metrics in all:

\textbf{(1) MMD.D:} maximum mean discrepancy (MMD) of degree distribution between the generated graphs and the test graphs.

\textbf{(2) MMD.C:} maximum mean discrepancy (MMD) of clustering coefficient between the generated graphs and the test graphs.

\textbf{(3) MMD.S:} maximum mean discrepancy (MMD) of spectrum between the generated graphs and the test graphs~\cite{efficient}.

\textbf{(4) MMD.O:} maximum mean discrepancy (MMD) of node orbit counts between the generated graphs and the test graphs~\cite{you2018graphrnn}.

\textbf{(5) $\alpha$:} The power-law exponent of the graph degree distribution.

\textbf{(6) $D_{k}$:} The Kolmogorov-Smirnov distance between the degree distributions of the generated and test graphs.

\textbf{(7) Valid:} Research demonstrates that degree distributions in complex networks are typically characterized by a power-law exponent $\alpha \in [2, 3]$~\citep{clauset2009power}. Accordingly, we define the validity measure for a graph as the proportion of graphs meeting the criteria $D_{k} < 0.1$ and $\alpha \in [2, 3]$. Set \( k_{\text{min}} = 2 \) for the uniform calculation of the power-law fitness of both undirected and directed graphs.

\textbf{(8) GEM:} To quantify the level of structural alignment for the expanded graph, we establish the Graph Expansion Metric (GEM). 
Firstly, for the negative indicator MMD metrics, we utilize the transformation \(1 - \frac{1}{1+e^{metric}}\), which maps the metrics to a range between 0 and 1. We then calculate the average of MMD and Valid metrics as GEM.

\paragraph{Experiment Settings}

Specifically, we sample a network dataset based on publication timelines to create our evaluation dataset. Since we only crawl for timestamp information of the CiteSeer and Cora datasets, these two datasets are used for our experimental evaluation. Following the timeline of graph evolution, we partition the network dataset into training and testing sets.  
At a designated time point \( t \), we filter the citation network using node timestamp information to obtain \( G_{<t} \), which includes all nodes and edges prior to \( t \). We sample small subgraphs from \( G_{<t} \) to create a training set for deep learning methods and to generate the seed graph for GAG. The training set consists of sampled subgraphs with sizes ranging from 64 to 512 nodes, resulting in a train set comprising 160 subgraphs and validation sets comprising 32 subgraphs.
For the test set, we filter the citation network for nodes and edges after \( t \), denoted as \( G_{>t} \). From \( G_{>t} \), we sample large subgraphs of 1,000 nodes, resulting in a test set comprising 20 subgraphs. We focus on whether the expanded graph structure exhibit power law characteristics typical of real-world network structures.

The existing graph generation methods have demonstrated promising results in generating small graphs, including works such as \cite{vignac2023digress}, \cite{martinkus2022spectre}, and \cite{you2018graphrnn}. However, they have not explored the generation of out-of-distribution graph sizes, and the generated graph sizes are limited. To compare with traditional graph generation models, we need to select those methods that can expand beyond the training set graph sizes and efficiently generate large graphs.
For rule-based graph generation methods, we set hyperparameters to ensure that the average degree of the expanded graph matches that of the seed graph. For deep learning-based graph generation methods, we adhere to the hyperparameters specified in the original papers. All hyperparameter details are provided in Table~\ref{table:hyper_params}.

\paragraph{Ablation on Seed Graph Size}

\begin{figure}[htbp]
  \centering
  {\includegraphics[width=\linewidth]{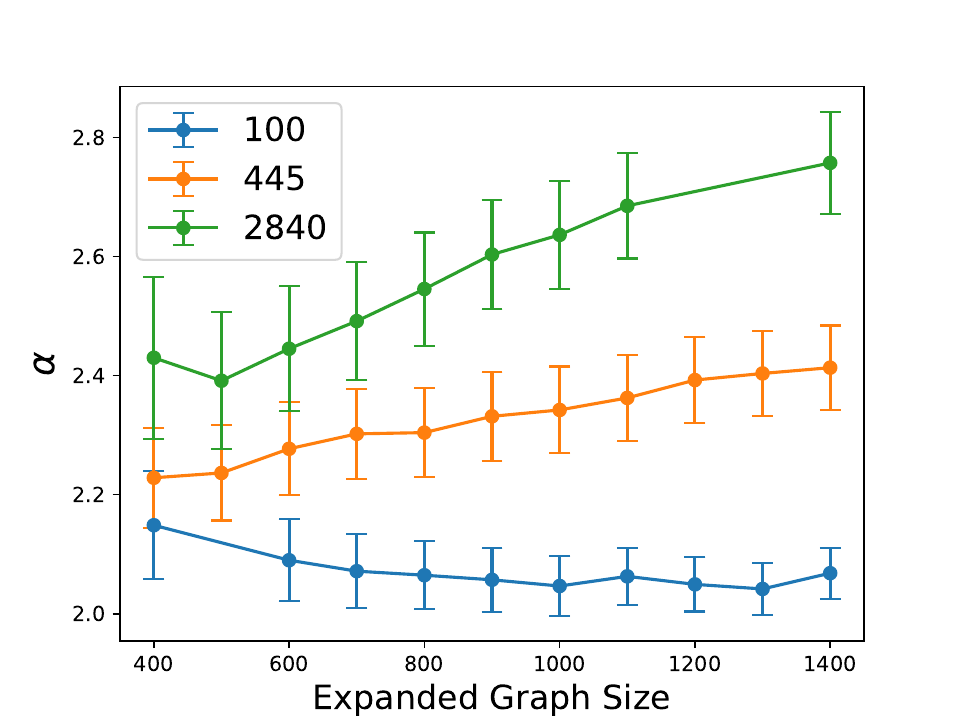}}
  \caption{We present the results for seed graph sizes of 100, 445, and 2840, with the sizes of the expanded graphs plotted against their corresponding \(\alpha\) values. Note that we only plot valid data \(D_k < 0.1\).}
  \label{fig:ab_seed_graph}
\end{figure}

Additionally, we aim to investigate whether the size of the seed graph affects the validity of the final generated graph structure. To this end, we utilize the GAG to perform graph expansion on seed graphs of varying sizes. We plot the number of nodes in the expanded graphs against the corresponding values of \(\alpha\). As shown in Figure~\ref{fig:ab_seed_graph}, it is evident that larger seed graphs result in expanded graphs exhibiting higher values of \(\alpha\). Furthermore, as the size of the expanded graphs increases, the \(\alpha\) values gradually stabilize. This indicates that the GAG is capable of effectively and reasonably expanding graphs across different seed graph sizes.

\paragraph{Supplementary Experiments}
\label{appendix:time}

To further demonstrate the reliability of the GAG framework, in addition to seed graph from CiteSeer, we crawl for the necessary node attributes to enrich text attributes for Cora~\cite{CiteGraph_Sen_08} to constitute the seed graph.
Following the experimental setup outlined in the paper, we designate Cora as the seed graph and similarly compared it with existing graph generation models.

As shown in Table \ref{table:compare_structure_cora}, the expanded graphs generated by GAG adhere to a power-law distribution, with $\alpha = 2.1$. 
For MMD metrics, since GAG doesn't strictly enforce explicit graph structure constraints on the agents, its performance is not significantly better than other models, but it achieves comparable results.
For the Valid metric, apart from GAG and the Barabási-Albert Model, the only existing deep-learning graph generation model capable of capturing the power-law distribution is L-PPGN. This demonstrates that L-PPGN is indeed capable of extrapolating to out-of-distribution graphs when trained on graphs that encompass the power-law distribution property. However, the unstable performance of L-PPGN across different datasets also highlights its sensitivity to the quality of the training dataset.
In contrast, GAG, through human behavior simulation, can reliably generate graph structures that adhere to real-world network characteristics from seed graph of varying sizes and quality. This illustrates not only the potential of LLM-based Agents in simulating human behavior but also underscores the reliability of the GAG framework.

\section{Textual Feature Alignment}
\label{section:ablation_text}

\subsection{Experiment Settings}
For the LLM backbone, we have chosen the open-source model of Llama-3-70B\cite{llama3modelcard} for textual feature alignment experiment. We configure the complete Citeseer graph as seed graph, and expand the paper citation graph to 5000 nodes. 
For baselines, we construct graphs with eight distinct relationships between graph structures and textual features as follows:

\textbf{(1) F.shuffle:} This process randomly selects node-wise textual features from seed graph to create the node features of expanded graph.

\textbf{(2) S.shuffle:} In this step, we randomly shuffle the edges of seed graph, thereby disrupting the graph structure while maintaining the number of edges consistent with \( \mathcal{V}^s \).

\textbf{(3) SF.shuffle:} This refers to a combination of both F.shuffle and S.shuffle.

\textbf{(4) BWR.L:} Graph strcuture generated by \cite{diamant2023improving}. We give the LLM Citeseer as corpus and use it to generate textual features.

\textbf{(5) L-PPGN.L:} Graph strcuture generated by \cite{bergmeister2024efficient}. We give the LLM Citeseer as corpus and use it to generate textual features.

\textbf{(6) BiGG.L:} Graph strcuture generated by \cite{dai2020scalable}. We give the LLM Citeseer as corpus and use it to generate textual features.

\textbf{(7) BiGG.L:} Graph strcuture generated by \cite{dai2020scalable}. We give the LLM Citeseer as corpus and use it to generate textual features.

\textbf{(8) GraphMaker.L:} Graph strcuture generated by \cite{li2024graphmaker}. We give the LLM Citeseer as corpus and use it to generate textual features.

\textbf{(9) GAG:} Graphs that are generated by the GAG framework.

\subsection{Ablation Study of LLM}
To further illustrate the effectiveness of the generated graphs, we conduct an ablation study on the LLM for agent setup. We select four LLMs for this study: GPT-3.5-turbo, GPT-4o-mini \cite{openai2023gpt4} as top-ranking closed-source LLMs, and  Llama-3-70B\cite{llama3modelcard} and Qwen2-72B\cite{yang2024qwen2technicalreport} as top-ranking open-source LLMs.

\begin{table*}[ht]
  \caption{Ablation Study of LLM in GAG. Performance Comparison for Node Classification}
  \label{tab:performance_comparison_nc_ablation}
  \centering
  \begin{tabular}{rrrrrr}
    \toprule
    % \multirow{2}{*}{GNN} 
    \multirow{2}{*}{} & & \multicolumn{4}{c}{$\Delta{ACC} \downarrow$} \\
    \cmidrule{3-6}
    GNN & LLM & SF.random & F.random & S.random & GAG \\
    \midrule
    GAT  
    & LLAMA. & $13.40_{\pm 3.52}$ & $24.11_{\pm 2.84}$ & $18.85_{\pm 2.55}$ & \textbf{2.26$_{\pm 1.19}$} \\
    & GPT-3. & $12.30_{\pm 2.47}$ & $22.24_{\pm 2.86}$ & $16.69_{\pm 2.46}$ &  \textbf{2.29$_{\pm 1.84}$} \\
    & GPT-4o. & $12.03_{\pm 1.17}$ & $21.73_{\pm 2.91}$ & $15.57_{\pm 2.84}$ & \textbf{2.80$_{\pm 0.88}$} \\
    & Qwen2. & $11.12_{\pm 2.53}$ & $15.84_{\pm 3.34}$ & $24.51_{\pm 2.28}$ & \textbf{10.50$_{\pm 1.94}$} \\
    GCN 
    & LLAMA. & $15.04_{\pm 1.99}$ & $23.20_{\pm 2.25}$ & $21.77_{\pm 2.04}$ & \textbf{3.61$_{\pm 1.32}$} \\
    & GPT-3. & $13.55_{\pm 2.19}$ & $23.08_{\pm 1.39}$ & $20.58_{\pm 1.44}$ & \textbf{4.34$_{\pm 2.01}$} \\
    & GPT-4o. & $14.56_{\pm 1.84}$ & $22.07_{\pm 2.25}$ & $20.82_{\pm 2.45}$ & \textbf{4.09$_{\pm 1.75}$} \\
    & Qwen2. & $12.56_{\pm 1.65}$ & $15.37_{\pm 2.69}$ & $26.95_{\pm 2.16}$ & \textbf{10.05$_{\pm 1.87}$} \\

    GCN2 
    & LLAMA. & $9.58_{\pm 0.88}$ & $9.33_{\pm 1.36}$ & $2.24_{\pm 1.58}$ & \textbf{0.49$_{\pm 1.50}$} \\
    & GPT-3. & $9.07_{\pm 1.10}$ & $9.04_{\pm 1.80}$ & $1.43_{\pm 1.67}$ & \textbf{0.39$_{\pm 1.67}$} \\
    & GPT-4o. & $8.89_{\pm 1.44}$ & $9.64_{\pm 1.10}$ & $1.19_{\pm 1.64}$ & \textbf{0.50$_{\pm 1.36}$} \\
    & Qwen2. & $9.78_{\pm 1.48}$ & \textbf{9.19$_{\pm 1.44}$} & $11.40_{\pm 1.78}$ & $9.39_{\pm 0.93}$ \\

    GraphSage 
    & LLAMA. & $7.00_{\pm 0.92}$ & $7.13_{\pm 1.73}$ & $3.15_{\pm 1.84}$ & \textbf{0.09$_{\pm 1.74}$} \\
    & GPT-3. & $6.72_{\pm 1.41}$ & $6.39_{\pm 0.92}$ & $2.35_{\pm 1.60}$ & \textbf{0.81$_{\pm 1.67}$} \\
    & GPT-4o. & $6.97_{\pm 0.92}$ & $6.46_{\pm 1.61}$ & $2.05_{\pm 2.31}$ & \textbf{1.70$_{\pm 2.28}$} \\
    & Qwen2. & \textbf{6.45$_{\pm 1.33}$} & $7.14_{\pm 1.68}$ & $12.73_{\pm 2.26}$ & $9.72_{\pm 2.43}$ \\
    \bottomrule
  \end{tabular}
  
\end{table*}

As shown in Table \ref{tab:performance_comparison_nc_ablation}, the LLM-based agents built on LLaMA2, GPT-3.5-turbo, and GPT-4o-mini are capable of generating graphs that maintain the node and structural characteristics of the seed graph, thus ensuring effective performance transfer in downstream tasks. In contrast, Qwen-2 based agents do not guarantee performance transfer. We believe this is related to the ability of LLMs to emulate human behavior; Qwen-2 based agents fail to exhibit human-like creative behavior, resulting in less coherent generated graphs.

\section{Scalability of GAG}
\label{appendix:section_scalibity}

GAG demonstrates the ability to generate text-attributed dynamic graphs at scales exceeding those of existing graph generation models. Most current methods are limited to producing graphs with up to 5,000 nodes, while specialized models designed for sparse large graphs, such as \citet{dai2020scalable}, are constrained to generating simple grid-structured graphs with a maximum of 100,000 nodes. Similarly, the GraphMaker model \citep{li2024graphmaker} considers graphs with 13,000 nodes to be large-scale. In contrast, our GAG framework is capable of generating graphs with up to 100,000 nodes that exhibit intricate small-world structures, without imposing sparsity assumptions. Moreover, GAG produces graphs that are both dynamic and text-attributed—a substantial advancement over existing approaches. While some models can generate either dynamic or attributed graphs, none match GAG’s ability to support the generation of text-attributed dynamic graphs at such scales, marking a key contribution of our work. To further clarify the scalability of the GAG framework, we have included an updated comparison in Table~\ref{table:graph_scale_compare}:

\begin{table*}[htbp]
    \caption{Comparsion of generated graphs by GAG against existing graph generation methods.}
    \label{table:graph_scale_compare}
    \centering
    \resizebox{\textwidth}{!}{%
    \begin{tabular}{l|llllll}
    \toprule
    \textbf{Model} & \textbf{Method} & \textbf{Text-Attributed} & \textbf{Attributed} & \textbf{Temporal} & \textbf{Scale(Nodes)} & \textbf{Year} \\
    \midrule
    GAG & Simulation & \cmark & \cmark & \cmark & 100000 & 2024 \\ 
    GraphMaker~\citep{li2024graphmaker} & Diffusion & & \cmark & & 13000 & 2024 \\ 
    L-PPGN~\citep{bergmeister2024efficient} & Diffusion & & \cmark & & 5037 & 2024 \\ 
    EDP-GNN~\citep{pmlr-v108-niu20a} & Diffusion & & \cmark & & <100 & 2020 \\ 
    MOOD~\citep{lee2023exploring} & Diffusion & & \cmark & & <100 & 2022 \\ 
    GDSS~\citep{jo2022score} & Diffusion & & \cmark & & <100 & 2022 \\
    DiGress~\citep{vignacdigress} & Diffusion & & \cmark & & <100 & 2021 \\
    Bwr-GraphRNN~\citep{diamant2023improving} & AR & & & & 5037 & 2023 \\ 
    BIGG~\citep{dai2020scalable} & AR & & & & 100000 & 2020 \\ 
    GRAN~\citep{efficient} & AR & & & & 5037 & 2019 \\ 
    GraphRNN~\citep{you2018graphrnn} & AR & & & & 2025 & 2018 \\
    MolecularRNN~\citep{popova2019molecularrnn} & AR & &\cmark & & <100 & 2019 \\
    GRAM~\citep{kawai2019scalable} & AR & & & & 500 & 2021 \\
    DeepGDL~\citep{khodayar2019deep} & AR & &  & & 14430 & 2019 \\
    LFM~\citep{podda2020deep} & AR & & & & <100 & 2020 \\
    STGG~\citep{ahn2021spanning} & AR & & & & <100 & 2021 \\
    MDVAE~\citep{du2022interpretable} & VAE & & \cmark & & <100 & 2022 \\
    D-MolVAE~\citep{du2022small} & VAE & & & & <100 & 2022 \\
    GraphVAE~\citep{simonovsky2018graphvae} & VAE & & \cmark & & <100 & 2018 \\
    % D2G2~\citep{zhang2021disentangled} & VAE & & \cmark & \cmark & 2500 & 2022 \\
    STGD-VAE~\citep{du2022disentangled} & VAE & & \cmark & \cmark & 2500 & 2022 \\
    CGVAE~\citep{liu2018constrained} & VAE & & \cmark & & <100 & 2018 \\ 
    JT-VAE~\citep{jin2018junction} & VAE & & \cmark & & <100 & 2018 \\ 
    GraphNVP~\citep{madhawa2019graphnvp} & NF & & \cmark & & <100 & 2019 \\ 
    GRF~\citep{honda2019graph} & NF & & \cmark & & <100 & 2019 \\ 
    MoFlow~\citep{zang2020moflow} & NF & & \cmark & & <100 & 2020 \\ 
    GraphAF~\citep{shi2020graphaf} & AR+NF & & \cmark & & <100 & 2019 \\ 
    GraphDF~\citep{luo2021graphdf} & NF & & \cmark & & <100 & 2021 \\ 
    MolGAN~\citep{de2018molgan} & GAN & & \cmark & & <100 & 2018 \\
    Mol-CycleGAN~\citep{maziarka2020mol} & GAN & & \cmark & & <100 & 2020 \\ 
    GraphEBM\citep{liu2021graphebm} & EBM & & \cmark & & <100 & 2021 \\
    \bottomrule
    \end{tabular}
    }
\end{table*}

\begin{table}[H]
  \caption{The time cost (\textit{min}) of agents for generating one interaction data with $P = 24$.}
  \label{table:time_cost_agent}
  \centering
  \begin{tabular}{rlll}
    \toprule
    N & SC & TC & SoC \\
    \midrule
    5 & 0.2700 & 0.0150 & 0.0120 \\
    10 & 0.2300 & 0.0112 & 0.0119 \\
    20 & 0.1700 & 0.0052 & 0.0119 \\
    40 & 0.0910 & 0.0054 & 0.0060 \\
    \midrule
    5$\rightarrow$40 & $\downarrow$\textbf{66.3\%} 
    & $\downarrow$\textbf{64.0\%} & $\downarrow$\textbf{50.0\%} \\
    \bottomrule
  \end{tabular}
\end{table}

\begin{table}[H]
  \caption{The time cost (\textit{hour}) for one round of simulation for generating the large-scale graphs.}
  \label{table:time_cost}
  \centering
  \begin{tabular}{rlll}
    \toprule
     & N & P & T \\
    \midrule
    SC & 5.01e+03 & 10 & 0.46h \\
    TC & 3.91e+03 & 10 & 0.30h \\
    SoC & 9.97e+04 & 48 & 11h \\
    \bottomrule
  \end{tabular}
\end{table}

To further demonstrate the excellent scalability of the GAG framework, we conduct time measurements across various simulation scenarios. The tests are carried out on a computing machine equipped with 96 CPU cores and 376 GB of memory. For model inference, we utilized LLAMA-3-70B as the backbone LLM and employed the vLLM framework~\cite{kwon2023efficient}, running on a setup of four A-800 GPUs. 
As shown in Table \ref{table:time_cost_agent}, when \( P \) is held constant, the time to generate one interaction data decreases as $N$ increases. The most significant time reduction observed in the SC simulation, where agents are grouped by paper authorship, which maximizes the efficiency of parrallel acceleration.  

For generating the large-scale graphs listed in Table~\ref{table:graph_structure_metrics}, we carry out SC simulation experiments for 200 rounds, TC simulation experiments for 33 rounds, and SoC simulation experiments for 10 rounds. For each scenario, we measured the total simulation time and computed the average simulation time per round. These multi-round simulations provide a robust measure of the computational efficiency of the GAG framework. 
The results, summarized in Table \ref{table:time_cost}, demonstrate GAG's capability to handle simulations with varying scales of agents.
For simulation experiments with Thousands of agents, the average simulation time per round is 0.46 hours for SC experiments and 0.30 hours for TC experiments; For simulation experiments with Hundreds of thousands of agents, the average simulation time per round is 11 hours for SoC.

\section{Ablation Study of S-RAG}
% \section{S-RAG Algorithm}

To investigate whether the hyperparameter settings of S-RAG affect the generated network structure. We conduct ablation experiments on these hyperparameters. Given the variations in graph generation scenarios, we conduct the ablation experiments under the SoC simulation. We run equal number of simulation rounds within the GAG to generate graphs.

In this section, we add a graph structure metric for measuring the proportion of the largest connect component within the network. We define the largest connect component of graph as $LCC$, so the proportion of $LCC$ within the network is $|LCC|/|V|$. This aids us in comprehending the graph evolution progress.

\subsection{Recall Stage}
In recall stage, the only hyperparameter is the number of searched items: $N_{r}$. Since the final number of documents interacting with the LLM is limited to $N_{r}$, we change $N_{r}$ and evaluate its impact on network structure.

\begin{table*}[h]
  \centering
  \caption{Ablation Study of $N_{r}$. The value of $N_{r}$ is proportional to $\bar{k}$ of the generated network.}
  \label{tab:ablation_N_fine}
  \begin{tabular}{ll|llllll}
    \toprule
    % \multirow{2}{*}{Network} & & \multicolumn{6}{c}{Network Structural Characteristics}
    % \\ \cmidrule(lr){3-8} 
    Network & $N_{r}$ & \( |\mathcal{V}^s| \)  &\( |\mathcal{E}| \)  & \( \bar{cc} \)  &\( r \) & $|LCC|/|V|$ \\
    \midrule
    \multirow{4}{*}{Action}& 3 & 9.47e+02 & 1.92e+03 & 0.07 & 0.10 & 0.03 \\
    & 5 & 9.36e+02 & 2.20e+03 & 0.09  & -0.05 & 0.02 \\
    & 10 & 9.58e+02 & 2.63e+03 & 0.09  & 0.02 & 0.05 \\
    & 20& 1.03e+03 & 3.03e+03 & 0.11 & -0.08 & 0.16 \\
    \midrule \\
    \multirow{4}{*}{Follow}& 3& 7.42e+02 & 1.27e+04 & 0.83  & -0.08 & 0.29 \\
    & 5 & 7.39e+02 & 1.24e+04 & 0.81  & -0.06 & 0.44 \\
    & 10& 7.39e+02 & 1.29e+04 & 0.80  & -0.06 & 0.51 \\
    & 20& 8.91e+02 & 3.83e+04 & 0.82  & -0.18 & 1.00 \\
    \midrule \\
    \multirow{4}{*}{Friend} & 3& 7.42e+02 & 5.96e+03 & 0.88 & -0.13 & 0.25 \\
    & 5& 7.39e+02 & 5.76e+03 & 0.87  & -0.10 & 0.22 \\
    & 10& 7.39e+02 & 5.94e+03 & 0.89 & -0.10 & 0.24 \\
    & 20& 8.91e+02 & 1.83e+04 & 0.87 & -0.10 & 0.45 \\
    \bottomrule
    \end{tabular}  
\end{table*}

As shown in Table \ref{tab:ablation_N_fine}, we keep all other search parameters constant while varying the size of \( N_{r} \). It can be observed that as \( N_{r} \) increases, $\bar{k}$ of generated network also exhibits an upward trend. This trend is particularly pronounced in the follow network.

\subsection{Reranking Stage}

To maximum the effectiveness of searched items, we implement the ReRanking stage in S-RAG. Initially, coarse ranking is performed to sort the searched items by their creator agent. 
Focusing on the \textit{core} lable of creator. Subsequently, fine ranking is conducted based on the agent's individual preferences. We conduct an ablation study to explore the impact of different levels of personalization in ReRanking stage. And eventually its impact on the network structure.

We focus on the hyperparameters in the ReRanking stage, which mainly include:
(1) Hub rate: $|HUB|/|V|$.
(2) Attributes of $a_l$.

\paragraph{Coarse Ranking}
\label{appendix:coarse_ranking}

\begin{table*}[htbp]
  \centering
  \caption{Ablation Study of the hub rate ($|HUB|/|V|$). Higher hub rate contributes to the emergence of a large connected component.}
  \label{tab:ablation_hub} 
  \begin{tabular}{ll|llllll}
  \toprule
  Network & $|HUB|/|V|$ & \( |\mathcal{V}^s| \)  &\( |\mathcal{E}^s| \)  & \( \bar{cc} \)  &\( r \) & $|LCC|/|V|$ \\
  \midrule
  \multirow{3}{*}{Action} & 0.00 & 9.78e+02 & 2.64e+03 & 0.09  & -0.05 & 0.13 \\
   & 0.10 & 1.02e+03 & 2.58e+03 & 0.09  & -0.07 & 0.10 \\
   & 0.20 & 1.03e+03 & 3.03e+03 & 0.11  & -0.08 & 0.16 \\
  \midrule \\
  \multirow{3}{*}{Follow} & 0.00 & 7.79e+02 & 3.00e+04 & 0.84 & 0.02 & 0.63 \\
   & 0.10 & 7.82e+02 & 3.04e+04 & 0.84 & 0.05 & 0.63 \\
   & 0.20 & 8.91e+02 & 3.83e+04 & 0.82 & -0.18 & 1.00 \\
  \midrule \\
  \multirow{3}{*}{Friend} & 0.00 & 7.79e+02 & 1.45e+04 & 0.89 & 0.21 & 0.27 \\
  & 0.10 & 7.82e+02 & 1.47e+04 & 0.88 & 0.29 & 0.43 \\
  & 0.20 & 8.91e+02 & 1.83e+04 & 0.87 & -0.10 & 0.45 \\
  \bottomrule
  \end{tabular}
\end{table*}

As shown in Table \ref{tab:ablation_hub}, an increase in the proportion of core users correlates with an upward trend in the proportion of the largest connected component within the network. Since the proportion of core users is increased, the likelihood of core users being searchable by general users is also increased, thereby fostering preferential attachment in the network. Eventually, the proportion of the largest connected component within the network is increased.

\paragraph{Fine Ranking}
\label{appendix:ablation_filter}
To improving search algorithms based on personal preferences of agents, we design various filter items in fine ranking process, which are tailored to different simulation scenarios. The number of filter items is $N_{f}$. Within the SoC simulation, filter items include:
(1) Follow: Assesses whether the content of the document is posted by an agent that the current agent follows.
(2) Friend: Assesses whether the content of the document is sent by an agent that is a friend of the current agent.
(3) Topic: Assesses whether the content of the document is related to a topic that the current agent is interested in. 

As illustrated in Table \ref{tab:ablation_filter}, \( \bar{cc} \) of network increases as $N_{f}$ increases. Additionally, the impact level of different filter items is as follows: friend $>$ topic $>$ follow.

\begin{table*}[htbp]
  \caption{Ablation Study of the fillter items used in fine ranking process.}
  \label{tab:ablation_filter}
  \centering
  \begin{tabular}{llll|lllll}
    \toprule
    & \multicolumn{3}{c}{Fillter Items} 
    & \multicolumn{5}{c}{Network Structural Characteristics}
    \\
    \cmidrule(lr){2-4} \cmidrule(lr){5-9} 
    % \cline{2-4} \cline{5-11}
    Network & follow & topic & friend & \( |\mathcal{V}^s| \)  &\( |\mathcal{E}^s| \)  & \( \bar{cc} \) &\( r \) & $|LCC|/|V|$ \\
    \midrule
    % \multirow{4}{*}{Action} & - & - & - & - & 1.12e+03 & 2.69e+03 & 0.15 & 0.21 & -0.09 & 0.06 \\
    \multirow{4}{*}{Action}& \cmark & - &  -  & 6.51e+02 & 1.88e+03 & 0.11 & 0.01 & 0.19 \\
    & - & \cmark &  - & 6.32e+02 & 1.82e+03 & 0.09  & -0.01 & 0.15 \\
    % & - & - & \cmark & - & 1.02e+03 & 2.29e+03 & 0.04 & 0.08 & -0.05 & 0.04 \\
    & - & - & \cmark & 1.02e+03 & 2.78e+03 & 0.09 & -0.05 & 0.12 \\
    & \cmark & \cmark & \cmark & 1.03e+03 & 3.03e+03 & 0.11 & -0.08 & 0.16 \\
    \midrule \\
    % \multirow{4}{*}{Follow} & - & - & - & - &  8.95e+02 & 7.65e+05 & 1.00 & 1.00 & 0.99 & 0.98 \\
    \multirow{4}{*}{Follow}  & \cmark & - & -  & 6.51e+02 & 2.65e+04 & 0.78 & -0.19 & 0.92 \\
    & - & \cmark & - & 5.63e+02 & 2.03e+04 & 0.79 & 0.16 & 0.64 \\
    % & - & - & \cmark & - &  9.31e+02 & 8.60e+04 & 0.95 & 1.00 & 0.95 & 0.29 \\
    & - & - & \cmark &  7.70e+02 & 2.97e+04 & 0.84  & 0.19 & 0.69 \\
    & \cmark & \cmark & \cmark & 8.91e+02 & 3.83e+04 & 0.82 & -0.18 & 1.00 \\
    \midrule \\
    % \multirow{4}{*}{Friend} & - & - & - & - & 8.95e+02 & 3.83e+05 & 1.00 & 1.00 & 1.00 & 0.98 \\
    \multirow{4}{*}{Friend} & \cmark & -  & -  &  6.51e+02 & 1.26e+04 & 0.83 & -0.13 & 0.46 \\
    & - & \cmark & - &  5.63e+02 & 9.74e+03 & 0.83 & 0.52 & 0.28 \\
    % & - & - & \cmark & - &  9.31e+02 & 4.26e+04 & 1.00 & 1.00 & 1.00 & 0.29 \\
    & -  & - & \cmark &   7.70e+02 & 1.44e+04 & 0.89 & 0.21 & 0.31 \\
    & \cmark & \cmark & \cmark & 8.91e+02 & 1.83e+04 & 0.87 & -0.10 & 0.45 \\
    \bottomrule
    \end{tabular}  
\end{table*}

\section{Human Interface Control}

\begin{figure*}[htbp]
  \centering
  \rotatebox{270}{\includegraphics[height=\linewidth]{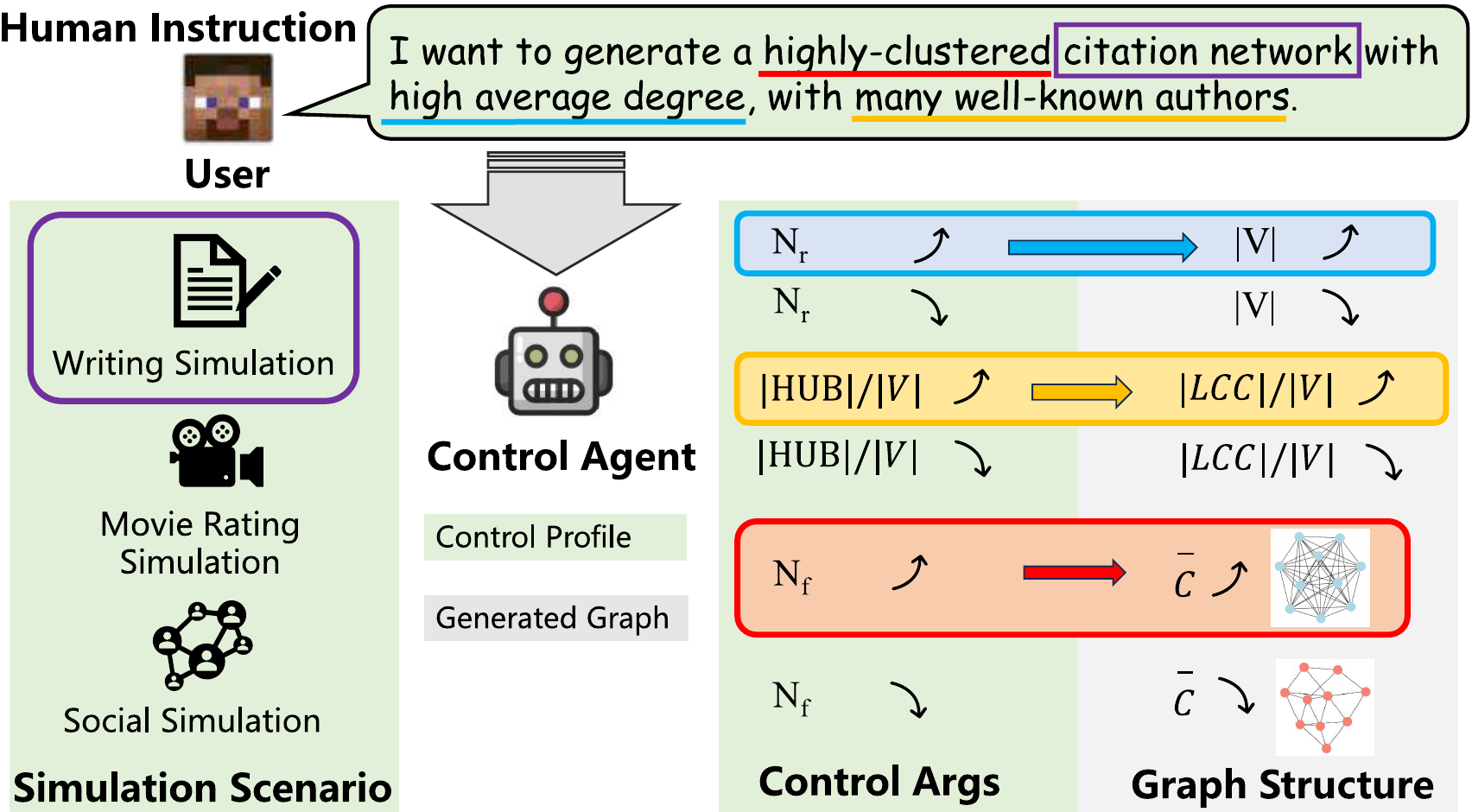}}
  \caption{An illustration of the Control Agent in GAG Framework.}
  \label{fig:Interface_control}
\end{figure*}

Previous work on employing LLMs for graph generation typically relied on predefined network structure features or a set of example networks \cite{yao2024exploring}. Similarly, after understanding the reasons behind different structural characteristics of networks within GAG, we aim to enable users to control the entire simulation process by inputting prompts. This will guide and influence the various structural features of the final network.

To achieve this, we establish a control agent that accepts instruction frm users. Control agent transfers the instruction to a control profile for managing the simulation process of GAG.
As shown in Fig. \ref{fig:Interface_control},specifically, the hub rate controls the proportion of recommended core users, subsequently affecting the ratio of hub nodes in the network. The parameter \( N_{r} \) determines the number of items recommended by the system, influencing the overall degree distribution. Additionally, parameter \( N_{f} \) dictate the number of filter items in the ReRanking stage, impacting the network's clustering coefficient. Furthermore, the overall simulation time is adjusted by the number of agents $N$ per simulation round.

\subsection{Case Study}

Since GAG employs human behavior simulation for network generation, the process of connecting each network node to others closely mirrors real-world scenarios. This alignment enables a clear and interpretable understanding of the network evolution process. To demonstrate the interpretability of our graph generation method, we present a case study using the SC simulation scenario.
\begin{figure}[htbp]
  \centering
  {\includegraphics[width=\linewidth]{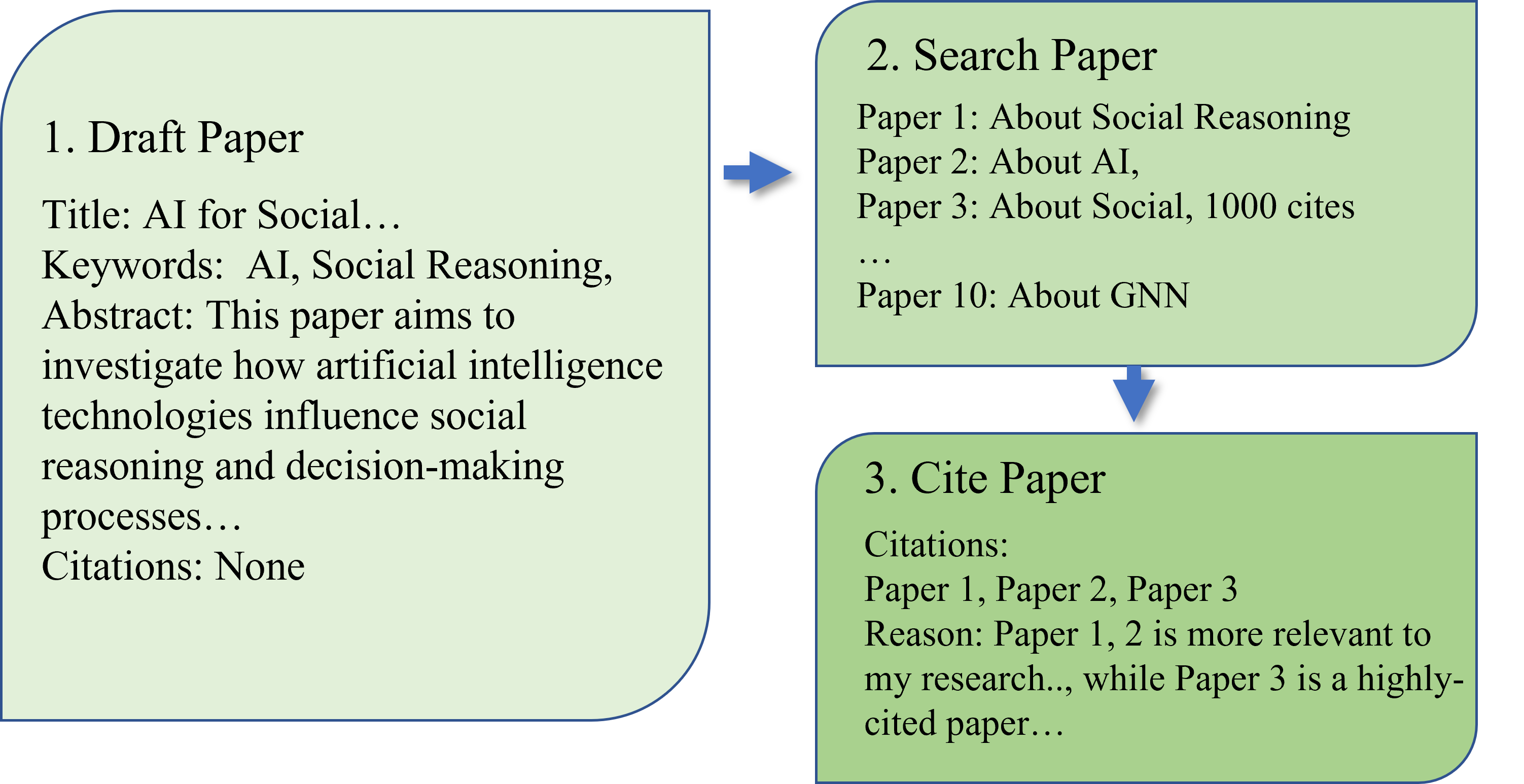}}
  \caption{An illustration of the citation network evolution with LLM-based Agents.}
  \label{fig:case_ca}
\end{figure}

 As illustrated in Figure \ref{fig:case_ca}, the formation of a citation network involves three primary steps. First, LLM-based agents collaboratively generate a paper draft through interaction and cooperation. Next, the agents search the corpus of stored papers within the environment to identify literature relevant to their research interests. For instance, query terms may correspond to research domains such as AI or social sciences.

Finally, after completing the search, the agents select and cite papers pertinent to their draft, providing explicit justifications for each citation. As illustrated in Figure \ref{fig:case_ca}, examples of such justifications include citing a paper due to its high citation count or its direct relevance to the research topic.
Each citation edge in the network thus directly corresponds to an agent's citation action, offering a behavior-driven perspective on the graph construction process. This approach ensures that the graph generation process is inherently interpretable.

\clearpage
% \section{Prompts}
% \label{appendix:prompts}

\begin{table*}[ht]
  \caption{The action prompt template and respond example in SC.}
  \label{prompt:paper_writing}
  \begin{tcolorbox}[colback=blue!5, 
  colframe=black,
  ]
  
  \textbf{Action Prompt Template}

  \textbf{Agent Profile:} $<$Agent Profile$>$

  \textbf{Agent Memory:} $<$Agent Memory$>$

  \textbf{Environment Observation}: Partial environmental feedback provided by S-RAG. The searched items are formed as textual representation with item prompt template.

  \textbf{Human Instruction}: 
      A papar should include the following attributes:  

        title: The title should be concise yet descriptive, providing a clear indication of the paper's topic and scope. This can be different from your topic, It is relatively accurate and clear. 

        keywords: These are specific terms or phrases that encapsulate the core topics of your paper. Keywords make your paper searchable within academic databases and help readers quickly understand the paper's focus areas. 

        abstract: The abstract is a brief summary of your research paper. It should provide an overview of the research question, methodology, results, and conclusions.  

        citations: A list of the paper names you want to cite. 

      \dots

      Now write a version of your paper and cite the papers you need to cite.
 
      {\rule{\linewidth}{1pt}\par\vspace{6pt}}

    \textbf{Structured Respond Example}
    
    \textbf{Creation Action}: Actor-Item Edge: Creation. Item Node: NULL.

    \textbf{Reference Action}: Actor-Item Edge: Reference. Item Node: Academic paper.

  \end{tcolorbox}
  \end{table*}

\begin{table*}[ht]
  \caption{The action prompt template and respond example in TC.}
  \label{prompt:movie_rating}
  \begin{tcolorbox}[colback=blue!5, 
  colframe=black,
  ]
  \textbf{Action Prompt Template}

  \textbf{Agent Profile:} $<$Agent Profile$>$

  \textbf{Agent Memory:} $<$Agent Memory$>$

  \textbf{Environment Observation}: Partial environmental feedback provided by S-RAG. The searched items are formed as textual representation with item prompt template.

  \textbf{Human Instruction}: 
    You should give your rating scores to the movies \dots

  {\rule{\linewidth}{1pt}\par\vspace{6pt}}
  
  \textbf{Structured Respond Example}

  \textbf{Movie Rating Action}: Actor-Item Edge: Rating. Item Node: NULL.

  \end{tcolorbox}
  \end{table*}

\begin{table*}[ht]
  \caption{The action prompt template and respond example in SoC.}
  \label{prompt:tweet_sending}
  \begin{tcolorbox}[colback=blue!5, 
  colframe=black,
  ]

  \textbf{Action Prompt Template}

  \textbf{Agent Profile:} $<$Agent Profile$>$

  \textbf{Agent Memory:} $<$Agent Memory$>$

  \textbf{Environment Observation}: Partial environmental feedback provided by S-RAG. The searched items are formed as textual representation with item prompt template.

  \textbf{Human Instruction}: 
 
  You can perform [Retweet/Reply/Tweet] action on these tweets. Additionally, you can follow the bloggers of these tweets:  

  Retweet: Retweet the tweet 

  Reply: Reply to the tweet 

  Tweet: Send a tweet 
  
  \dots
 
  {\rule{\linewidth}{1pt}\par\vspace{6pt}}

  \textbf{Structured Respond Example}

  \textbf{Retweet Action}: Actor-Item Edge: Retweet. Item Node: NULL.

  \textbf{Reply Action}: Actor-Item Edge: Reply. Item Node: NULL.

  \textbf{Follow Action}: Actor-Item Edge: Follow. Item Node: NULL.

  \textbf{Creation Action}: Actor-Item Edge: Tweet. Item Node: Tweets.

  \end{tcolorbox}
  \end{table*}

\begin{table*}[ht]
  \caption{The profile prompt template in SC.}
  \label{table:profile_prompt_sc}
  \begin{tcolorbox}[colback=blue!5, 
  colframe=black,
  ]
  I would like you to generate a series of random author's personal information.

These authors are interested in computer science, they are experts in various fields of CS.

I need you to give a list of author infos with the constraints for each attribute as follows:

  (1) Name: Author's name

  (2) Expertises: a list, The author's areas of expertises can be selected from the following areas:\{expertises list\}

  (3) Institution: The author's institution, you can choose whatever institution you want, just give me one institution name

  (4) Country: The author's country, you can choose whatever institution you want,just give me one country name corresponding to the institution

  (5) Topics: a list, The topics this author is interested in, can be selected from the following topics:\{topics list\}

  Here's some common used countrys you can infer to:

  \{countrys\}

  Please generate me a list of \{author num\} different authors, which can be loaded by eval function in python:

  [\{\{

  "name":"",

  "expertises":[],

  "institution":"",

  "country":"",

  "topics":[]

  \}\},

  ...,

  \{\{

  "name":"",

  "expertises":[],

  "institution":"",

  "country":"",

  "topics":[]

  \}\}]

  Now please generate:

  \end{tcolorbox}
  \end{table*}

  \begin{table*}[ht]
    \caption{The profile prompt template in TC.}
    \label{table:profile_prompt_tc}
    \begin{tcolorbox}[colback=blue!5, 
    colframe=black,
    ]
    Your task is to give me a list of watcher's profiles. Respond in this format:

[
  \{

  "gender": (F/M)

  "age":(the age of the watcher)

  "job":(the job of the watcher)

  \}
]

Respond:

    Now please generate:

    \end{tcolorbox}
    \end{table*}

  \begin{table*}[ht]
    \caption{The profile prompt template in SoC.}
    \label{table:profile_prompt_soc}
    \begin{tcolorbox}[colback=blue!5, 
    colframe=black,
    ]
    Your task is to give me a list of \{num added\} person's profiles for twitter users. Respond in this format:
[
\{\{
"user name": "(str;The name of this user)",
"user description":"(str;short and concise, a general description of this user, ordinary users or super \
large users and the topics this person interested in)"
\}\}
]

Now please generate:

    \end{tcolorbox}
    \end{table*}

\end{document}